%% file: main.tex
\documentclass{article}

\usepackage[utf8]{inputenc} 
\usepackage[T1]{fontenc}    
\usepackage{hyperref}       
\usepackage{url}            
\usepackage{booktabs}       
\usepackage{amsfonts}       
\usepackage{nicefrac}       
\usepackage{microtype}      

\usepackage{textcomp}

\usepackage{amsmath}
\usepackage{amsfonts}
\usepackage{amssymb}
\usepackage{amsthm}

\usepackage{graphicx}
\usepackage{rotating}
\usepackage{subcaption}
\usepackage[
textheight=9in,
textwidth=5.5in,
top=1in,
headheight=12pt,
headsep=25pt,
footskip=30pt]{geometry}

\usepackage{algorithm}
\usepackage{algorithmic}

\newcommand{\R}{\mathbb{R}}
\newcommand{\N}{\mathbb{N}}
\renewcommand{\P}{\mathbb{P}}
\renewcommand{\d}{\textnormal{d}}
\newcommand{\E}{\mathbb{E}}

\newcommand{\T}{\mathcal{T}}
\newcommand{\D}{\mathcal{D}}
\DeclareMathOperator*{\argmax}{\textnormal{argmax}}
\DeclareMathOperator*{\argmin}{\textnormal{argmin}}

\theoremstyle{definition}
\newtheorem{definition}{Definition}

\theoremstyle{plain}
\newtheorem{lemma}{Lemma}
\newtheorem{theorem}{Theorem}

\theoremstyle{remark}

\begin{document}
{

\include{main_paper}
}
\newpage
\appendix
\setcounter{lemma}{0}
\setcounter{theorem}{0}
\setcounter{definition}{0}
{

\include{supplement}
}
\end{document}

%% file: main_paper.tex
\title{Counterfactual Explanations of Concept Drift}

%

\author{%
  Fabian Hinder \\
  Cognitive Interaction Technology (CITEC)\\
  Bielefeld University\\
  Inspiration 1, D-33619 Bielefeld, Germany \\
  \texttt{fhinder@techfak.uni-bielefeld.de} \\
  \\
  Barbara Hammer \\
  Cognitive Interaction Technology (CITEC)\\
  Bielefeld University\\
  Inspiration 1, D-33619 Bielefeld, Germany \\
  \texttt{bhammer@techfak.uni-bielefeld.de} \\
}


\maketitle

\begin{abstract}
  The notion of concept drift refers to
the phenomenon that the distribution, which is underlying the observed data,
changes over time;
as a consequence machine learning models may become inaccurate and need adjustment. 
While there do exist methods to detect concept drift or to adjust models in the presence of observed drift, the question of \emph{explaining} drift has hardly been considered so far.
This problem is of importance, since it enables  an inspection of the most prominent
features where  drift manifests itself; hence it enables human understanding of the necessity of change and it increases acceptance of life-long learning models.
In this paper we present a novel technology, which characterizes concept drift in terms of the characteristic change of spatial features represented by typical examples based on counterfactual explanations.
We establish a formal definition of this problem, derive an efficient algorithmic solution based on counterfactual explanations,
and demonstrate its usefulness in several examples.
\end{abstract}

\section{Introduction}
\label{sec:intro}
One  fundamental assumption in classical machine learning is the fact that observed data are i.i.d.\ according to some unknown probability $\P_X$, i.e.\ the data generating process is stationary. 
Yet, this assumption is often violated 
in real world problems: models are subject to seasonal changes, changed demands of individual customers, ageing of sensors, etc. In such settings, 
life-long model adaptation rather than classical batch learning is required.
Since drift or covariate change 
is a major issue in real-world
applications,
many attempts were made to deal with this setting \cite{DBLP:journals/eswa/MelloVFB19,DBLP:journals/cim/DitzlerRAP15}.

Depending on the domain of data and application, the presence of drift is 
modelled in different ways.
As an example, covariate shift  refers to different marginal distributions of  training and test set  \cite{5376}. Learning for data streams extends this setting to an unlimited (but usually  countable) stream of observed data, mostly in supervised learning scenarios \cite{asurveyonconceptdriftadaption,DBLP:journals/tnn/ZambonAL18}.
Here one distinguishes between virtual and real drift, i.e.\ non-stationarity of the marginal distribution only or also the posterior. Learning technologies for such situations often rely on windowing techniques, and adapt the model based on the characteristics of the data in an observed time window. Active methods explicitly detect drift, while passive methods continuously adjust the model \cite{DBLP:journals/cim/DitzlerRAP15,DBLP:journals/kais/LosingHW18,JMLR:v19:18-251,DBLP:journals/connection/WangMCY19}. 

Interestingly, a majority of approaches deals with supervised scenarios, aiming for a small interleaved train-test error; this is accompanied by first approaches 
to identify particularly relevant features where drift occurs
\cite{DBLP:journals/corr/WebbLPG17}, and a large number of methods aims for a
detection of drift, an identification of change points in given data sets, or a characterization of overarching types of drift
\cite{Aminikhanghahi:2017:SMT:3086013.3086037,DBLP:journals/kais/GoldenbergW19}. 
However non of those methods aims for an \emph{explanation}\/ of the observed drift by means of a characterization of the observed change in an intuitive way. Unlike the vast literature on explainability of AI models \cite{ijcai2019-876,NIPS2019_9511,DBLP:conf/dsaa/GilpinBYBSK18,Gunningeaay7120}, only few approaches address explainability in the context of drift.
A first approach for explaining drift highlights
the features with most variance \cite{DBLP:journals/corr/WebbLPG17}; yet this approach is
restricted to an inspection of drift in single features. The purpose of our contribution is to provide a novel formalization how explain observed drift,
such that an informed monitoring of the underlying process becomes possible.
For this purpose, we characterize the underlying distribution in terms of typical representatives, and we 
describe drift by the evolution of these characteristic samples over time.
Besides a formal mathematical characterization of this objective, we provide an efficient algorithm to describe the form of drift and we show its usefulness in benchmarks. 

This paper is organized as follows: 
In the first part (sections~\ref{sec:setup} and \ref{sec:def}) we describe the setup of our problem and give a formal definition (see Definitions~\ref{def:i} and \ref{def:C}).
In section~\ref{sec:estimate} we derive an efficient algorithm as a realization of the problem.
In the second part we quantitatively evaluate the resulting algorithms and demonstrate their behavior in several benchmarks (see section~\ref{sec:experiments}).

\section{Problem Setup}
\label{sec:setup}

In the classical batch setup of machine learning one considers a generative process $P_X$, i.e.\ a probability measure, on $\R^d$. In this context one views the realizations of i.i.d.\ random variables $X_1,...,X_n \sim P_X$ as samples.
Depending on the objective, learning algorithms try to infer the data distribution based on these samples or, in the supervised setting, a  posterior distribution. We will only consider  distributions in general, this way subsuming the notion of both, real drift and virtual drift.

Many processes in real-world applications are online with data $x_i$ arriving consecutively as drawn from a possibly changing distribution, hence it is reasonable to incorporate temporal aspects. One prominent way to do so is to consider an index set $\T$, representing time, and a collection of probability measures $p_t$ on $\R^d$ indexed over $\T$, which describe the underlying probability at time point $t$ and which may change over time \cite{asurveyonconceptdriftadaption}. 
In the following we investigate the relationship of those $p_t$. Drift refers to the fact that $p_t$ is different for different time points $t$, i.e.\
\begin{align*}
    \exists t_0,t_1 \in \T : p_{t_0} \neq p_{t_1}.
\end{align*}
A relevant problem is to explain concept drift, i.e.\ characterize the difference of those pairs $p_t$. 
A typical use case  is the monitoring of processes. While drift detection technologies enable  automatic drift identification \cite{eddm,adwin, LSDD, hdddm,ddm,pagehinkley,Wald}, it is often unclear how to react to such drift, i.e.\ to decide whether a model change is due. This challenge is in general ill-posed and requires expert insight; hence an explanation would enable a human to initiate an appropriate reaction. 
A drift characterization is particularly demanding for high dimensional data or a lack of clear semantic features. 

In this contribution, we propose to describe the drift characteristics by contrasting suitable representatives of the underlying distributions \cite{molnar2019interpretable, Wachter2017CounterfactualEW}. 
Intuitively, we identify critical samples of the system, and we monitor their development
over time, such that the user can grasp  the characteristic changes as induced by the observed drift. 
This leads to the overall algorithmic scheme:
\begin{enumerate}
    \item Choose characteristic samples $(x_1,t_1),...,(x_n,t_n) \in \D$ that cover $\D$, where $\D$  denotes the set of observations / samples 
    (over data and time).
    \item For each sample $x_i$ find a corresponding $(x^{(i)}_j)_{j \in \T}$ such that $x^{(i)}_i = x_i$ and $(x^{(i)}_j,j) \in \D$ for all $j$, i.e.\ extend $x_i$ to 
    a time series of its corresponding points under drift.
    \item Present the evolution $(x^{(i)}_j)$, or its most relevant changes, respectively, to the user.
\end{enumerate}

In this intuitive form, however, this problem is still ill-posed. In the following, we formalize the notion of "characteristic points" for the distribution of $\D$ via optima of a characterizing function, and we define the problem of "correspondences" of samples within different time slices; these definitions will reflect our intuition and lead to efficient algorithmic solutions. 


\section{Characteristic Samples}
\label{sec:def}
To make the term "characteristic sample" formally tractable, we describe the process in terms of dependent random variables $X$ and $T$ representing data and time. This allows us to identify those values of $X$ that are "characteristic" for a given time and hence yields a notion of characteristic sample using information theoretic techniques.
To start with, we restrict ourselves to the case of discrete time, i.e.\ $\T \subset \N$, which is a particularly natural choice in the context of data streams or time series \cite{asurveyonconceptdriftadaption}. Even for continuous time, it is possible to find a meaningful discretization induced by change points by applying drift detection methods \cite{adwin,hdddm}. For simplicity, we assume finitely many time points, i.e.\ $\T = \{1,...,n\}$. This allows us to construct a pair of random variables $X$ and $T$, representing data and time respectively, which enable a  reconstruction of the original distributions by  the conditional distributions of $X$ given $T$, i.e.\ for $t \in \T$ it holds 
    $X | T = t \sim p_t$.
This corresponds to the joint distribution
\begin{align*}
    (T,X) \sim \frac{1}{|\T|} \sum_{t \in \T} \delta_t \times p_t,
\end{align*}
where $\delta_t$ denotes the Dirac-measure concentrated at $t \in \T$ and $P \times Q$ denotes the product measure. 
This notion has the side effect that, if we keep track of the associated time points of observations, i.e.\ we consider $(x_i,t_i)$ rather than just $x_i$, we may consider observations as i.i.d.\ realizations of $(X,T)$. In particular, we may apply well known analysis techniques from the batch setting.

\begin{figure}
    \centering
    \begin{minipage}[b]{0.45\textwidth}
    \includegraphics[width=\textwidth]{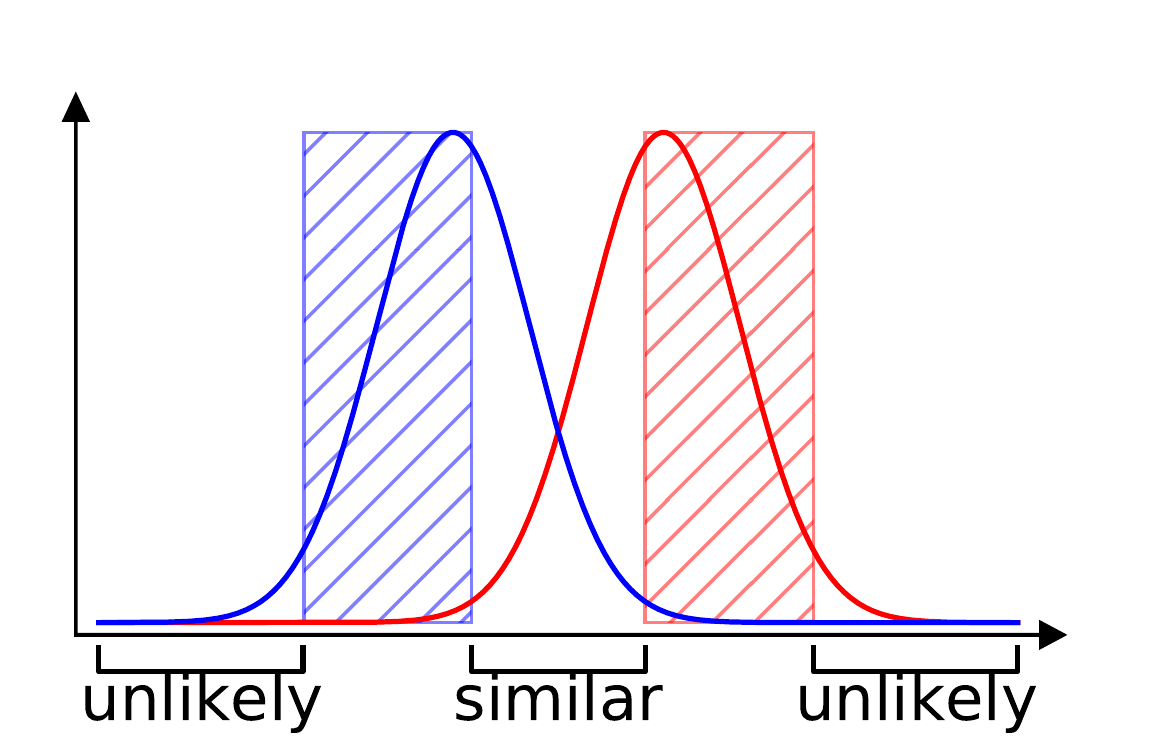}
    \subcaption{{Schematic illustration of two one-dimensional distributions (density; red and blue graphs) and their characteristic regions (hatched boxes).}
    \label{fig:comic:def_characteristic_1d}}
\end{minipage}
\hfill
\begin{minipage}[b]{0.45\textwidth}
    \centering
    \includegraphics[width=\textwidth]{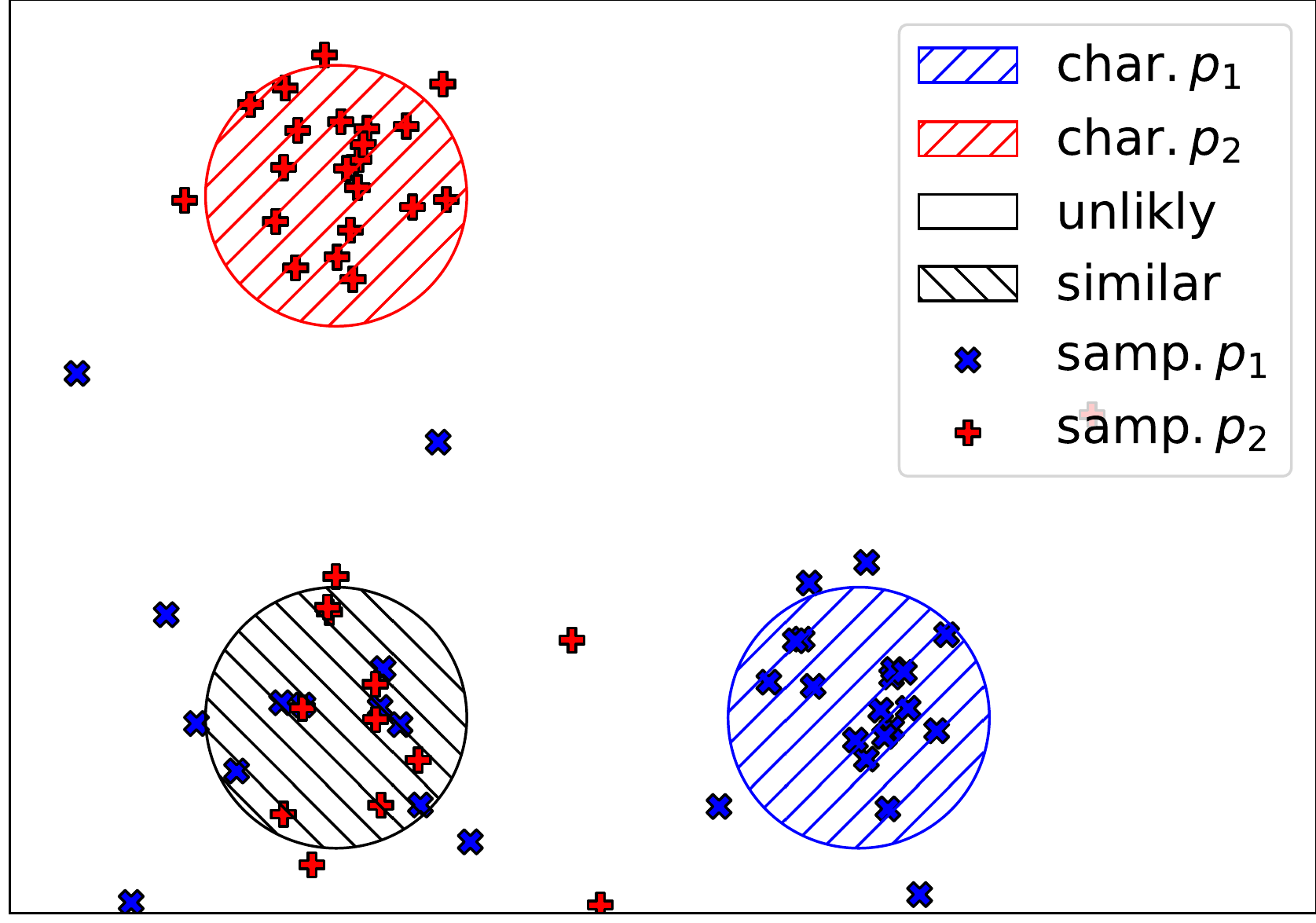}
    \subcaption{{Schematic illustration of two data sets (red crosses and blue X'es) and their characteristic and similar regions (red and blue res. black hatched).}
    \label{fig:comic:def_characteristic_2d}}
\end{minipage}
\caption{Schematic illustration of   characteristic samples.}
\label{fig:comic:def_characteristic}
\end{figure}

The term "characteristic" refers to two properties: the likelihood to observe such samples at all and the identifiability, which refers to the capability of identifying its origin, such as its generating latent variable, e.g.\ a certain point in time%
, this we quantify by means of entropy.
We illustrate this behaviour in Figure~\ref{fig:comic:def_characteristic}.
Here $X$, as defined above, is distributed according to a mixture model, where the origin is given by  the corresponding mixture component. Each of those components corresponds to a certain time point $t \in \T$. 
Informally, we say that an observation or  property identifies a certain time, if it only occurs during this time point.
By using Bayes' theorem we can characterize identifiability regarding $T$ for a given $X$ --  the probability that a certain data point $X=x$ was observed at $T=t$. Measuring its identifiability in terms of the entropy, we obtain the following definition:

\begin{definition}
\label{def:i}
The \emph{identifiability function}\/ induced by $p_t$ is defined as
\begin{align*}
    i(x) := 1-\frac{1}{\log |\T|}H(p_x)
\end{align*}
where $p_x$ is  induced by $f_x(t) := (\d p_t / \d \; |\T|^{-1} \sum_{t' \in \T} p_{t'})(x)$ over the uniform distribution on $\T$, where $\d \nu / \d \mu$ is the Radon-Nikodým density, $H(P) = -\sum_{t \in \T} P(t) \log P(t)$ denotes the entropy. 
\end{definition}

Obviously, $i$ has values in $[0,1]$. 
The identifiability function indicates the existence of drift as follows:

\begin{theorem}
\label{thm:lemma1}
$p_t$ has drift if and only if $\E[i(X)] \neq 0$.
\end{theorem}

Theorem~\ref{thm:lemma1} shows that $i$ captures important properties of $p_t$ regarding drift. It is important to notice that the identifyability function turns time characteristics into spatial properties: while drift is defined globally in the data space  and  locally in time, $i$ encodes drift locally in the data space and globally in time.  This will allow us to localize  drift, a phenomenon of time, in space, i.e.\ point towards spatial regions where drift manifests itself -- these can then serve as a starting point for an explanation of drift under the assumption that data points or features  have a semantic meaning. 


The identifiability function per se, however, does not take the overall probability into account. So  unlikely samples can be considered as identifying as long as they occur only at a single point in time. To overcome this problem, we extend $i$ to the characterizing function:

\begin{definition}
\label{def:C}
Let $\P_X$ denote the (density of) marginal distribution of $X$.
The \emph{characterizing function}\/
is defined as 
\begin{align*}
    C(x) := \P_X(x)i(x).
\end{align*}
We say that $x$ is a \emph{characteristic sample}\/ iff it is a (local) maximum of $C$.
\end{definition}

\begin{figure}
    \centering
    \includegraphics[width=0.8\textwidth]{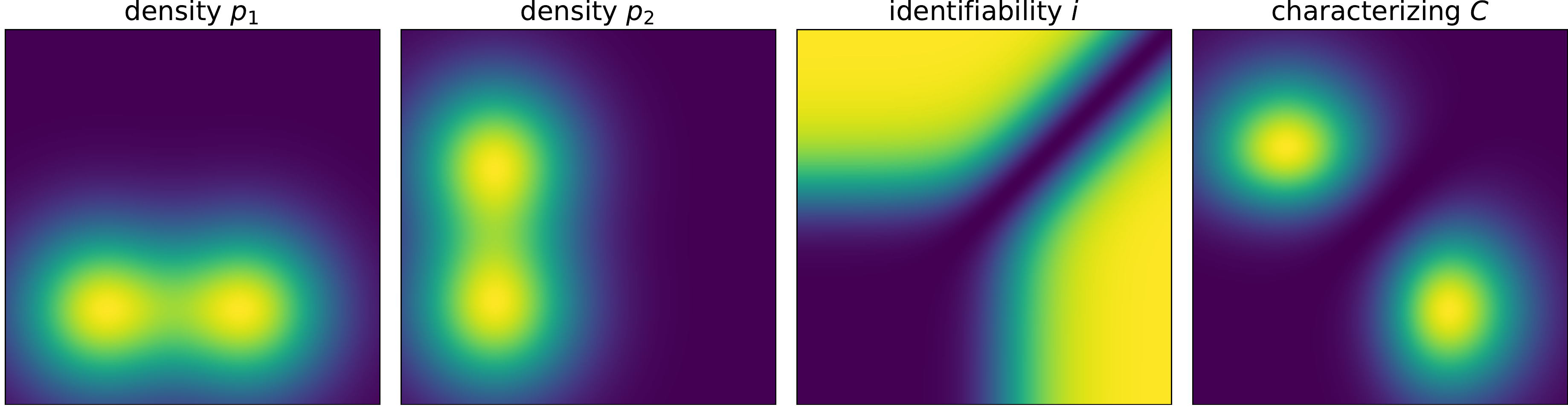}
    \caption{Head map of two distribution (also presented in Figure~\ref{fig:comic:def_characteristic_2d}) and their identifiability and characterizing map as defined in Definitions~\ref{def:i} and \ref{def:C} }
    \label{fig:comic:characteristic}
\end{figure}

In contrast to the identifiability function, the characterizing function also takes the likelihood of observing $x$ at any time into account. This reflects the idea, that a characteristic example is not only particularly pronounced with respect to other samples of another distribution, and hence identifiable, but also likely to be observed. We illustrate the behaviour of $i$ and $C$ in Figures~\ref{fig:comic:characteristic} and~\ref{fig:comic:def_characteristic_2d}. Obviously $C$ finds exactly those regions, which mark the presence of drift in the naive sense.

\subsection{Find Characteristic Samples given Data}
\label{sec:estimate}
We are interested in an efficient algorithm, which enables us to
 find characteristic samples from
 given data.
Unlike classical function optimization, we face the problem that  $C$ itself is unknown, and we cannot observe it directly. 
Rather, $C$  is given as a product of two components, each of which requires a  different estimation scheme. 
 We will rely on the strategy to estimate the identifiability function first. Then, we can reduce the problem to an estimation of a (weighted) density function, rather than estimating  $\P_X$ separately and then optimizing the  product $C$.

The problem of finding local maxima of a density function from given samples is a well studied problem, which can be addressed by prototype based clustering algorithms such as mean shift  \cite{MeanShift}, which identifies local maxima  based on kernel density estimators. Efficient deterministic counterparts  such as $k$-means often yield acceptable solutions \cite{Bish2}. 
Since $C$ constitutes a "weighted"  density function rather than a pure one,  we  rely on a weighted version of a prototype base clustering algorithm, which 
applies weighting/resampling  of samples according to the estimated identifiability function. The following theorem shows, that this procedure yields a valid estimate.

\begin{theorem}
\label{thm:density}
\newcommand{\A}{\mathcal{A}}
\newcommand{\B}{\mathcal{B}}
\newcommand{\C}{\mathcal{C}}
\renewcommand{\H}{\mathcal{H}}
\newcommand{\I}{\mathbb{I}}
\newcommand{\Xs}{\mathbf{X}}
Let $(\Omega,\A,\P)$ be a probability space, $(S, \B)$ a measure space and $X,X_1,X_2,...$ a sequence of $S$-valued, i.i.d. random variables. Let $f : S \to \R_{\geq 0}$ be a bounded, measurable map with $\E[f(X)] = 1$. Denote by $W(A) := \E[\I_A(X)f(X)]$ the $f$ weighted version of $\P_X$, where $\I$ denotes the indicator function. For every $n \in \N$ let $Y_1^{(n)},Y_2^{(n)},...$ be a sequences of independent $\{1,...,n\}$-valued random variables with $\P[Y_i^{(n)} = j|X_1,...,X_n] = f(X_j) / \sum_{k = 1}^n f(X_k)$ (or $1/n$ iff all $f(X_i) = 0$) for all $i = 1,2,...$ and $j = 1,...,n$. If $\C \subset \B$ is a Glivenko–Cantelli class of $\P_X$ then it holds
\begin{align*}
    W_{n,m} := \frac{1}{m} \sum_{i = 1}^m \delta_{X_{Y_i^{(n)}}} &\xrightarrow{m,n \to \infty} W 
\end{align*}
in $\Vert \cdot \Vert_\C$ almost surely, where we can take the limits in arbitrary order. 
\end{theorem}

Theorem~\ref{thm:density} implies that samples obtained by resampling from $\D$ according to $i$ are distributed according to the distribution associated to $C$, i.e.\ $A \mapsto \int_A C(x)\d x / \int C(x) \d x$. This induces an obvious algorithmic scheme, by applying prototype-based clustering to reweighted samples. This is particularly beneficial since some algorithms, like mean shift, do not scale well with the number of samples. It remains to find a suitable method to estimate $i$:
We need to estimate the probability of a certain time given a data point. Since we consider discrete time, this can be modeled as probabilistic classification problem which maps observations to a probability of the corresponding time, $h:X \to T$. Hence popular classification methods such as $k$-nearest neighbour, random forest, or Gaussian proceses can be used. We will evaluate the suitability of these methods in section~\ref{sec:exp:i}.

%

\section{Explanation via Examples: Counterfactual Explanations}
So far we have discussed the problem of finding characteristic samples, which can be modelled as probabilistic classification. This links the problem of explaining the difference between drifting distributions to the task of  explaining machine learning models by means of examples. One particularly prominent explanation in this context is offered by counterfactual explanations: these contrast samples by counterparts with minimum change of the appearance but different class label (see section~\ref{sec:setup}). First,
we shortly recapitulate  counterfactual explanations.

\subsection{Counterfactual Explanations}
Counterfactuals explain the decision of a  model  regarding a given sample by contrasting it with a similar one, which is classified differently \cite{Wachter2017CounterfactualEW}:

\begin{definition}
\newcommand{\X}{\mathcal{X}}
\newcommand{\C}{\mathcal{C}}
Let $h : \X \to \C$ be a classifier, $\ell : \C \times \C \to \R$ a loss function, and $d : \X \times \X \to \R$ a dissimilarity. For a constant $C>0$ and target class $y \in \C$ a counterfactual for a sample $x \in \X$ is defined as
\begin{align*}
\argmin_{x' \in \X} \ell(h(x'),y) + C d(x',x).
\end{align*}
\end{definition}
Hence a counterfactual 
of a given sample $x$ is an element $x' \in \mathcal{X}$ that is similar to $x$ but classified differently by $h$.
%
Common choices for $d$ include $p$-norms \(d(x,x') = \Vert x-x' \Vert_p^p = \sum_i |x_i-x_i'|^p \) or the Mahalanobis distance \(d(x,x') = (x-x')^T \Omega (x-x'),\)
with $\Omega$ as symmetric pdf matrix. 

As discussed in \cite{artelt2020convex,looveren2019interpretable} this initial definition  suffers from the problem that counterfactuals might be  implausible.
 To overcome this problem, the proposal \cite{looveren2019interpretable} suggests to allow only those samples that lie on the data manifold. This can be achieved by enforcing a lower threshold $\alpha > 0$ for the probability of counterfactuals
\begin{align*}
\argmin_{x' \in \mathcal{X}} \;&\;\ell(h(x'),y) + C d(x',x) \\ \text{s.t.} \:&\: \P_X(x') > \alpha
\end{align*}
In the work \cite{artelt2020convex}, $\P_X$ is chosen as mixture model, and approximated such that the optimizaton problem becomes a  convex problem for a number of popular models $h$.

\begin{algorithm}[!h]
   \caption{Explaining drift}
   \label{alg:alg1}
\begin{algorithmic}[1]
   \STATE {\bfseries Input:} {$S$ data stream}
   \STATE $\D \gets \emptyset; \D_0 \gets \emptyset; T \gets 1; \T \gets \emptyset$
   \WHILE{$\textsc{HasMoreSamples}(S)$}
   \STATE $x_\text{new} \gets \textsc{GetNextSample(S)}$
   \IF{$\textsc{HasDrift}(\D_0 \cup \{x_\text{new}\})$}
   \STATE $\D_0 \gets \emptyset; \T \gets \T \cup \{T\}; T \gets T+1$
   \STATE $h \gets \textsc{TrainProbabilisticClassifier}(\D)$
   \FORALL{$(x,t) \in \D$}
   \STATE $i_h[x] \gets 1-H(h(x))/\log|\T|$
   \ENDFOR
   \STATE $\D' \gets \textsc{ChooseRandomWeighted}(\D, i_h)$
   \STATE $\mathcal{C}_0 \gets \textsc{FindClusterprototypes}(\D')$
   \STATE $\mathcal{C} \gets \textsc{FindClosestPoint}(\mathcal{C}_0,\D)$
   \FORALL{$(x,t),(x',t') \in \mathcal{C} \times \D$}
   \STATE 
   $d[t'][x,x'] \gets \begin{cases}\textsc{Dist}(x,x') & t \neq t' \\ 0 & x = x'\\ \infty & \text{otherwise} \end{cases}$
   \ENDFOR
   \FORALL{$t \in \T$}
   \STATE 
   $A[t] \gets \textsc{AssignCounterfactual}(d[t])$
   \ENDFOR
   \STATE Present $\mathcal{C}$, $A$ to User
   \ENDIF
   \STATE $\D_0 \gets \D_0 \cup \{ x_\text{new} \}; \D \gets \D \cup \{ (x_\text{new},T) \}$
   \ENDWHILE
\end{algorithmic}
\end{algorithm}

\subsection{Explaining Drift by Means of Counterfactuals}
In section~\ref{sec:estimate} we connected the problem of identifying relevant information of observed drift to a probabilistic classification problem, mapping representative samples to their time of occurrence via $h$. This connection enables us to link the problem of understanding drift to the problem of explaining this  mapping by counterfactuals. We identify characteristic samples as local optima of $C$, as described above, and show how they contrast to similar points, as computed by counterfactuals, which are associated to a different time. 

Since we are interested in an overall explanation of the ongoing drift, we can also  restrict ourselves to
finding counterfactuals of $h$ within the set of given training samples, skipping the step of density estimation $\P_X$ to generate reasonable counterfactuals. It is advisable to coordinate the assignment of subsequent counterfactuals by minimizing the overall costs induced by the similarity matrix -- we refer to the resulting samples as \textit{associated samples}. 

The algorithmic scheme presented in section~\ref{sec:intro} gives rise to algorithm~\ref{alg:alg1}. The explaining routine is run if drift was detected.
Depending on the chosen sub algorithms (we use the Hungarian method, $k$-NN classifier, affinity propagation or $k$-means) we obtain a run time complexity of $\mathcal{O}(nm^2+m^2 \log m)+2\mathcal{O}(n^2)+\mathcal{O}(m)$, with $n$ the number of samples and $m$ the number of displayed representative samples for the processing of a drift event. Since $m \ll n$ we therefore obtain a run time complexity of $\mathcal{O}(n^2)$.


\begin{table}[b]
    \centering
    \caption{MSE for estimation identifiability function and final value of optimization of the identifiability function using different models/methods.
    First three data set are theoretical, encoding $d/n_{\text{GpC}}/n_\text{C}$ (dimension / complexity of distribution / complexity of component overlap). Estimation over 30 runs. Standard deviation is only shown if $\geq 0.01$. All results and details are given in the supplement.}
    \begin{tabular}{cccccc}
        & \multicolumn{2}{c}{ Estimation of $i$ (MSE)} & \multicolumn{3}{c}{Optimization of $i$ (mean value)} \\
        data set & $k$-NN & RF & $k$-M & AP & MS \\
        \hline
2/2/2&$0.01$&$0.08(\pm 0.03)$&$1.0(\pm 0.02)$&$0.99(\pm 0.05)$&$0.95(\pm 0.1)$\\
100/8/2&$0.01$&$0.2$&$1.0$&$1.0$&$1.0$\\
2/2/10&$0.06(\pm 0.01)$&$0.07$&$0.56(\pm 0.09)$&$0.62(\pm 0.09)$&$0.45(\pm 0.24)$\\
        \hline
        diabetes&$0.13(\pm 0.02)$&$0.10(\pm 0.01)$&$0.74(\pm 0.09)$&$0.77(\pm 0.14)$&$0.40(\pm 0.27)$\\ 
faces&$0.16(\pm 0.02)$&$0.14(\pm 0.02)$&$0.80(\pm 0.10)$&$0.82(\pm 0.10)$&$0.34(\pm 0.18)$\\ 
        \hline
    \end{tabular}
    \label{tab:quant}
\end{table}
\section{Experiments}
\label{sec:experiments}
In this section, we quantitatively evaluate the method. This includes the evaluation of the single components, and an application to realistic benchmark data sets.

\subsection{Quantitative Evaluation}
\label{sec:exp:i}
We evaluate the following components: Estimation of the identifiability map $i$, identification of characteristic samples, and plausibility of explanation for a known generating mechanism.
To reduce the complexity, we restrict ourselves to two time points, $\T = \{1,2\}$, since  multiple time points  can be dealt with by an iteration of this scenario.
We evaluate the estimation capabilities of different machine learning models -- $k$-nearest neighbour ($k$-NN), Gaussian process classification (GP with Matern-kernel), artificial neural network (ANN, 1-layer MLP) and random forest (RF) -- and prototype based clustering algorithms -- $k$-Means ($k$-M), affinity propagation (AP) and mean shift (MS). We evaluate on both theoretical data with known ground truth generated by mixture distributions,
as well as  common benchmark data sets for regression and classification for more realistic data, where its occurrence is induced by the output component. We present a part of the results in table~\ref{tab:quant}. Details are in the supplemental material.

\paragraph*{Evaluation of identifiability map} For the theoretical data,  we evaluate how a) dimensionality, b) complexity of distribution, and c) complexity of component overlap influence the model performance. As it turns out, the overlap is a crucial parameter, regardless of the chosen model. Further, $k$-NN is the least vulnerable method with best results,  random forests perform second best. 
For the benchmark data sets we found that $k$-NN performs quite well in most cases and is very similar to the random forest. The Gaussian process only works well on the regression data sets. 


%

\paragraph*{Evaluation of characteristic samples} 
We compared different prototype based clustering algorithms as regards their ability to identify representatives of $C(x)$. We applied the resampling scheme from section~\ref{sec:estimate} and also considered the weighted version of $k$-means as well as the standard version of $k$-means as baseline. It turns out that the resampling method performs best. Data parameters such as overlap and dimensionality have no significant influence. 
For the benchmark data sets we only evaluate the identifiability. We find that AP performs best, followed by $k$-means with resampling. 

\begin{figure}[!t]
    \centering
    \begin{minipage}[b]{0.325\textwidth}
    \includegraphics[width=\textwidth]{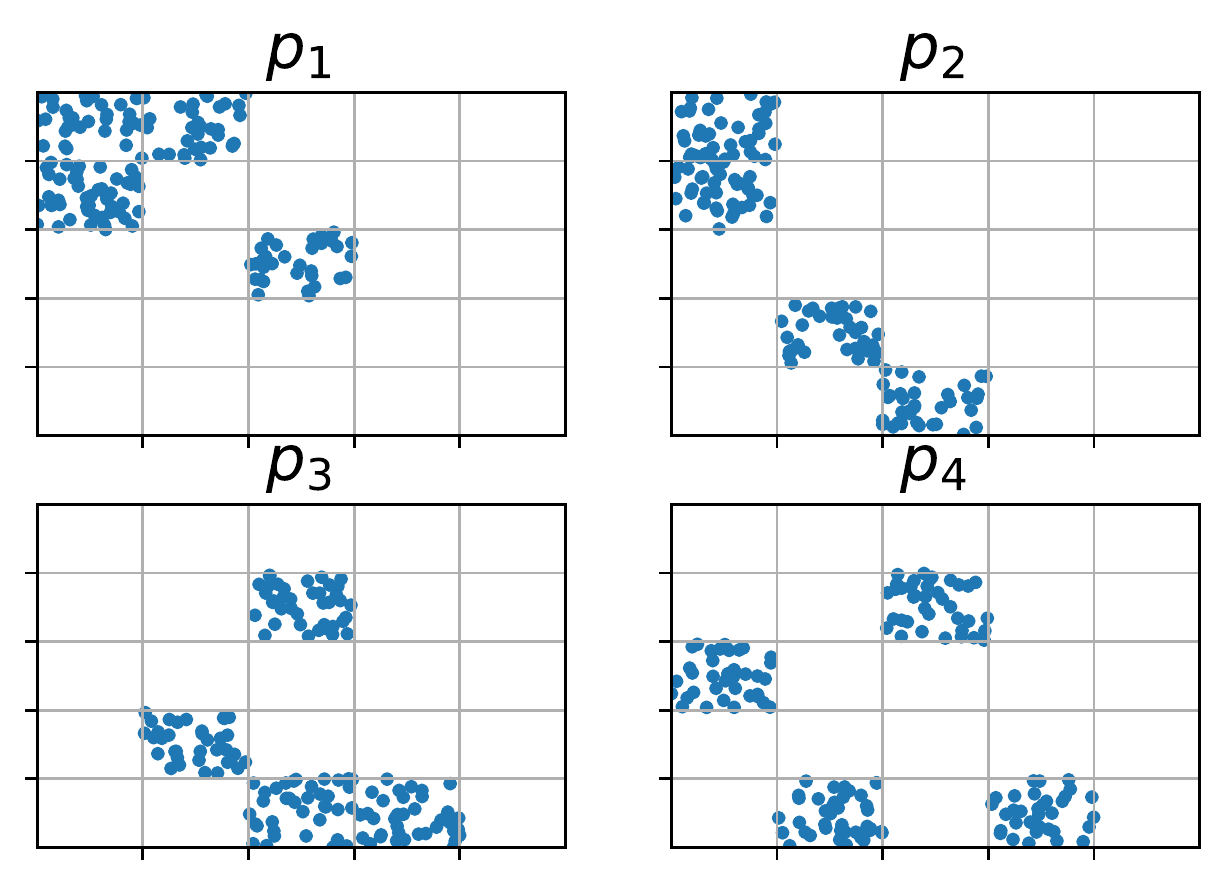}
    \subcaption{Evolving distribution with known components per time slot. 
    Distributions are given as mixtures of equally weighted uniform distributions.
    }
    \label{fig:eval:chess:overview}
    \end{minipage}
    \hfill
    \begin{minipage}[b]{0.325\textwidth}
    \centering
    \includegraphics[width=0.9\textwidth]{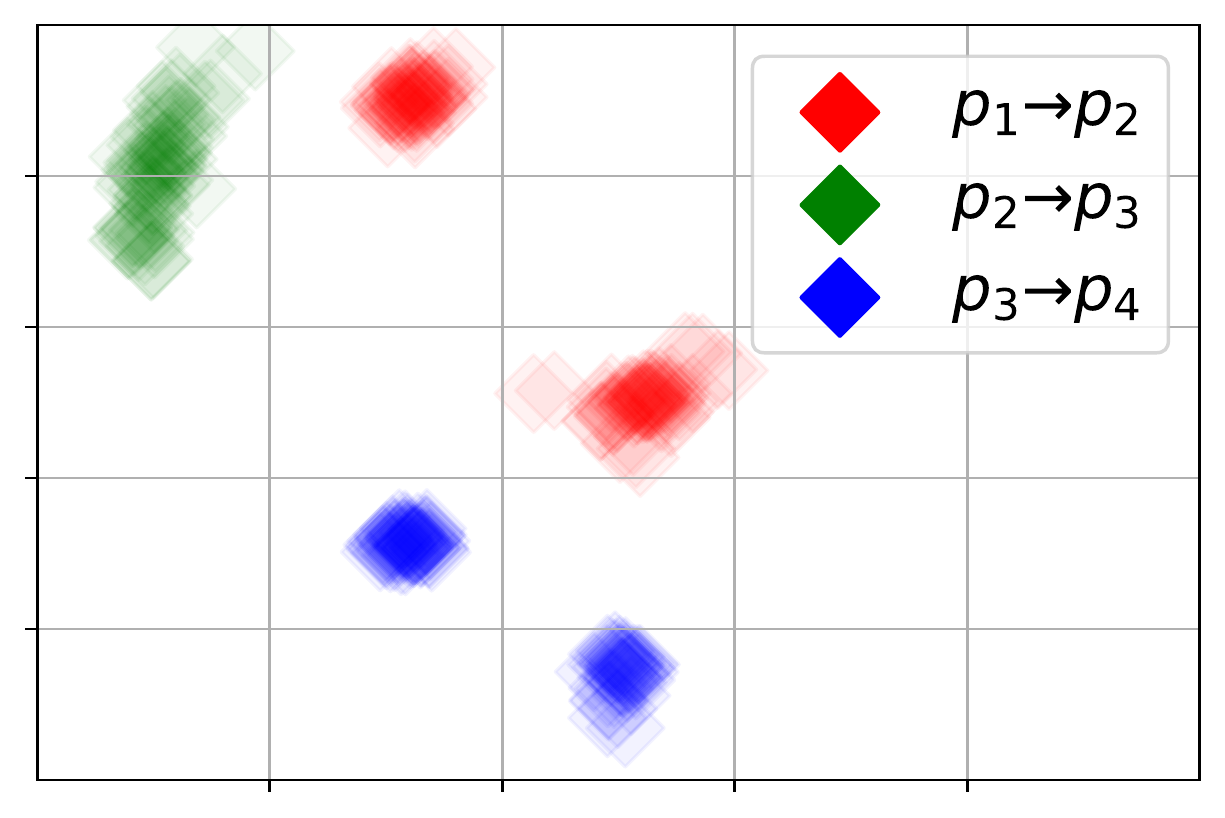}
    \subcaption{Visualization of the identified characteristic samples over 50 runs ($k$-nn \& mean shift). Underlying distributions are shown in Figure~\ref{fig:eval:chess:overview}. 
    }
    \label{fig:eval:chess:move}
    \end{minipage}
    \hfill
    \begin{minipage}[b]{0.325\textwidth}
    \includegraphics[width=\textwidth,height=3cm]{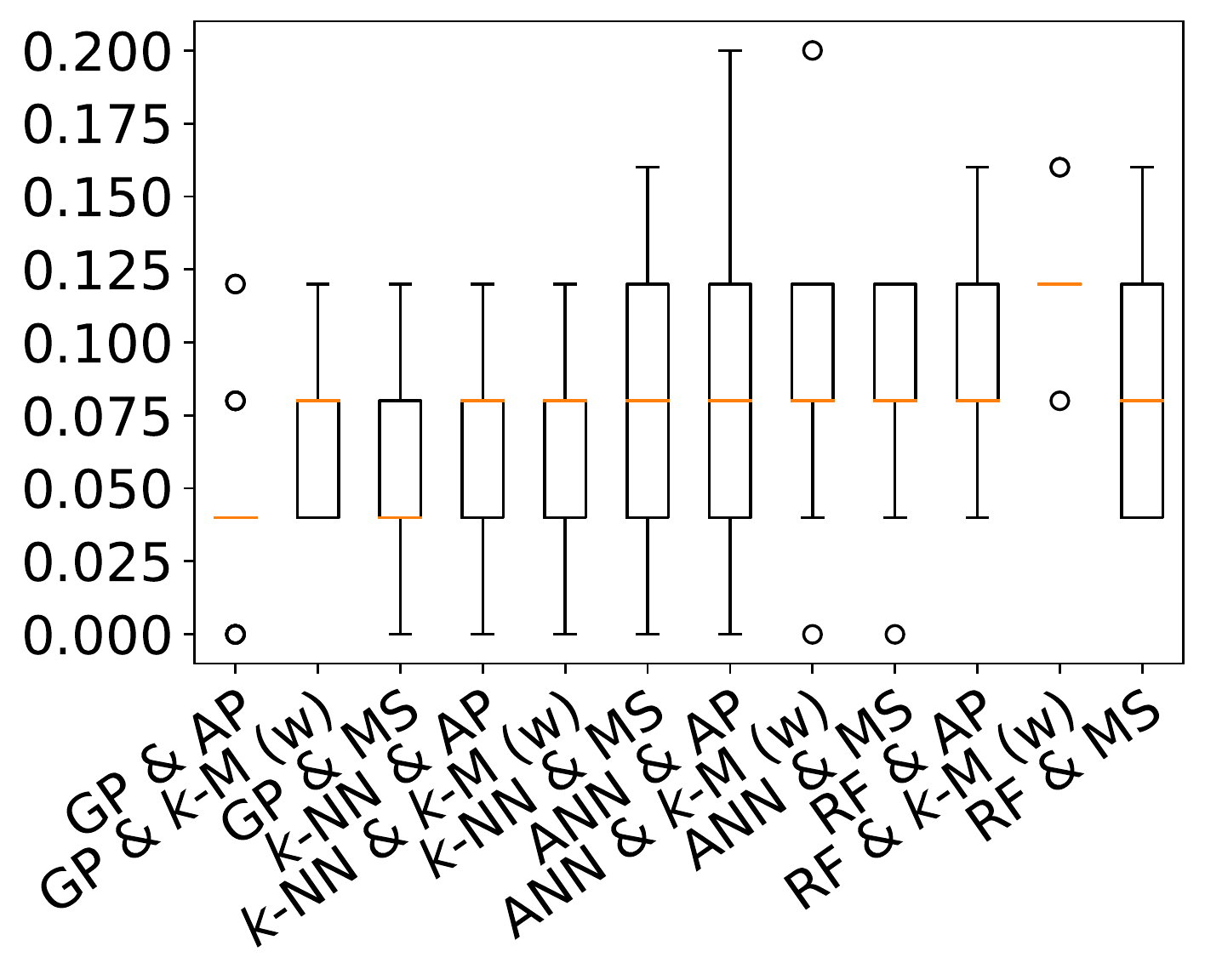}
    \subcaption{Performance of different instantiations of the algorithmic pipeline. Numbers refer to the mean percentage of misclassified cells.
    }
    \label{fig:eval:chess:performace}
    \end{minipage}
    \caption{Evaluation of the correct identification of spatial components provided by counterfactuals}
    \label{fig:eval:chess}
\end{figure}

\paragraph*{Evaluation of explainability} We evaluate the explainibility  by measuring the capability to detect vanishing of parts of the distribution. We generate a checkerboard data set (see Figure~\ref{fig:eval:chess:overview}) and evaluate the explanations as provided by the technology as regards its capability to identify parts which vanish/appear in subsequent time windows  (see Figure~\ref{fig:eval:chess:move}). 
A quantitative evaluation can be based on the number of incorrectly identified components, averaged over 30 runs, as shown in Figure~\ref{fig:eval:chess:performace}, using random distributions and $2\times 150$ samples. GP combined with AP 
performs best. 

\subsection{Explanation of Drift Data Sets}
\begin{figure}[!t]
    \centering
    \begin{minipage}[b]{0.35\textwidth}
    \includegraphics[width=\textwidth]{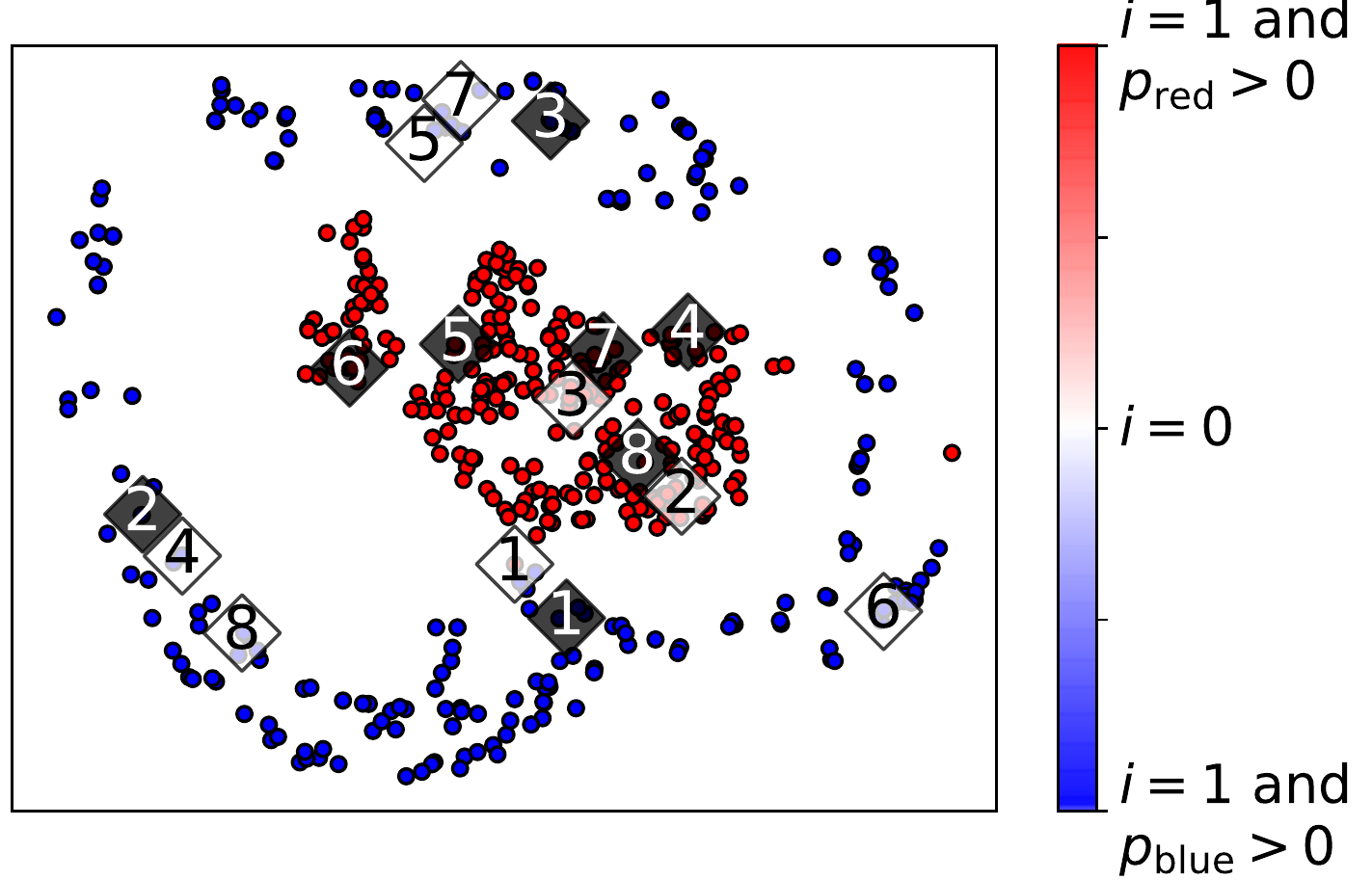}
    \subcaption{{Color represents origin (red/blue) and identifyability (satiation). Cards mark considered samples (black: characteristic sample; white: associated sample), projection via UMAP}\label{fig:exp:elec:overview}}
\end{minipage}
\hfill
\begin{minipage}[b]{0.62\textwidth}
    \centering
    \includegraphics[width=\textwidth]{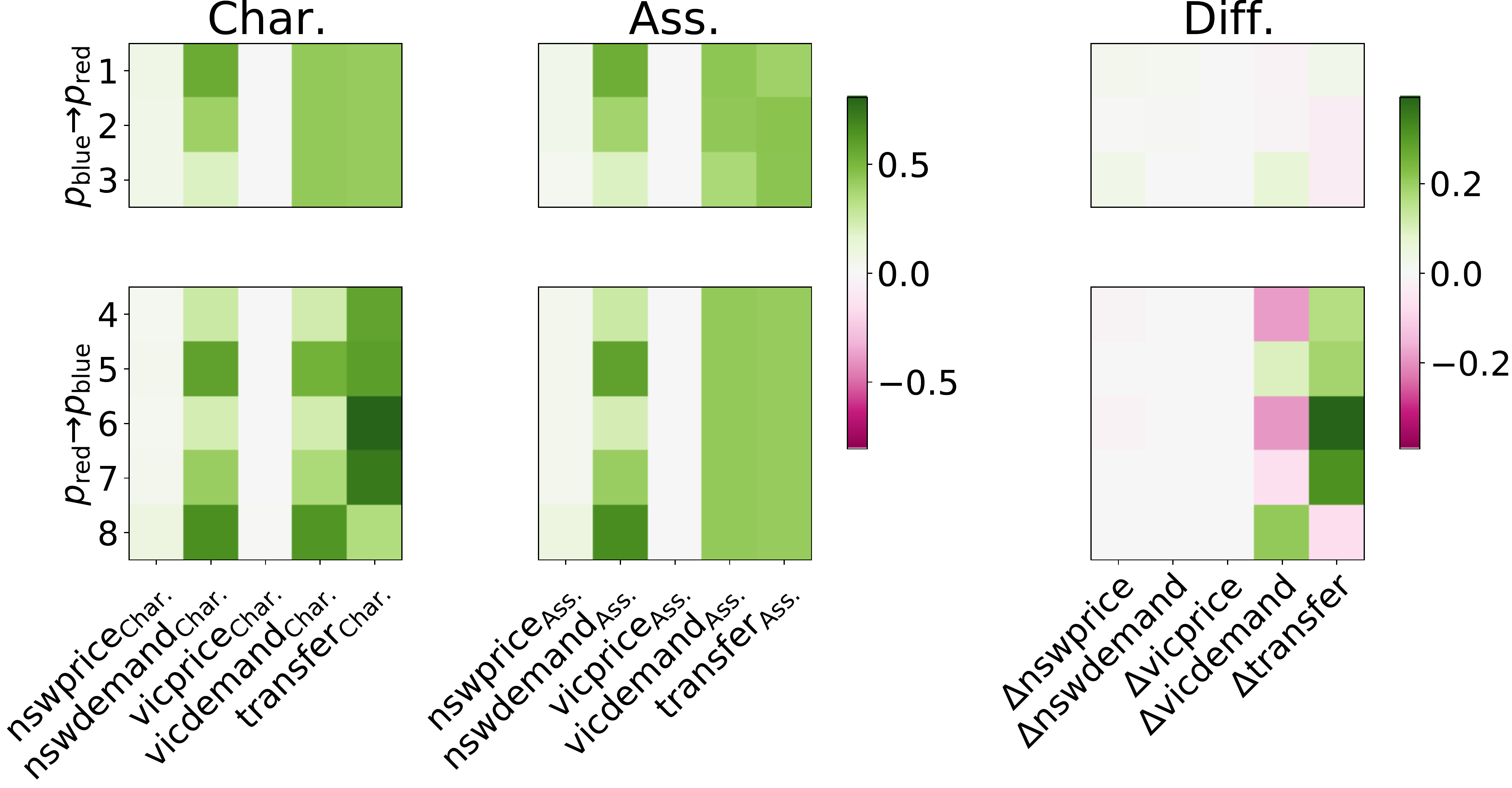}
    \subcaption{{Illustration of resulting pairs of samples: 
    Columns (left to right): characteristic sample, associated sample and difference of samples, Rows: from old to new (upper row), from new to old (lower row). 
    }\label{fig:exp:elec:detail}}
\end{minipage}
\caption{Our method applied to electricity market data set with split at 2nd of May 1997.}
\label{fig:exp:elec}
\end{figure}

\begin{figure}[b]
    \centering
    \begin{minipage}[b]{0.4875\textwidth}\centering
    \includegraphics[width=0.8\textwidth,height=3cm]{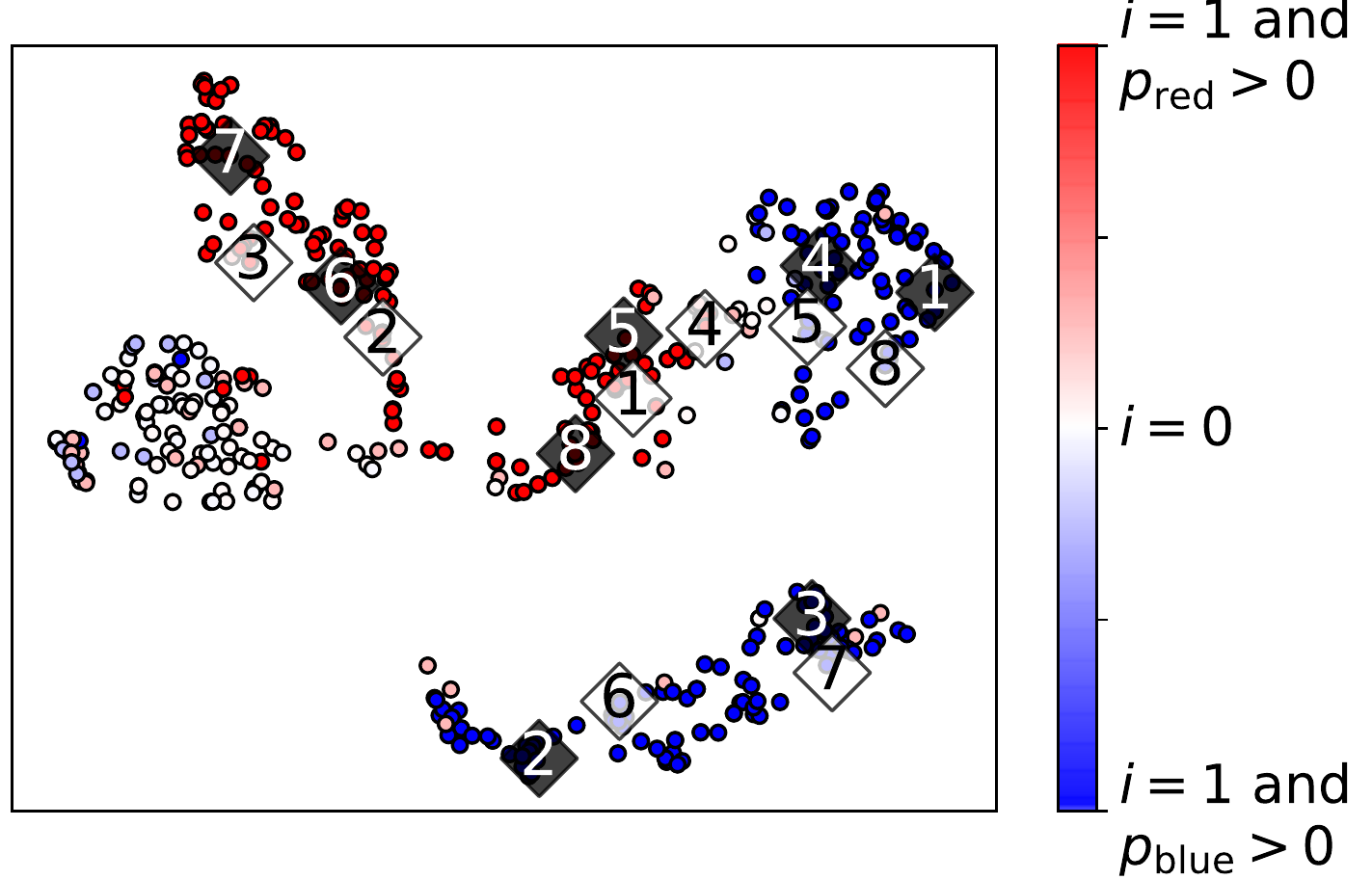}
    \subcaption{Overview image created using UMAP, same colors / markers as in Fig.~\ref{fig:exp:elec:overview}. 
    \label{fig:exp:mnist:overview}}
\end{minipage}
\hfill
\begin{minipage}[b]{0.4875\textwidth}
    \centering
    \includegraphics[width=\textwidth]{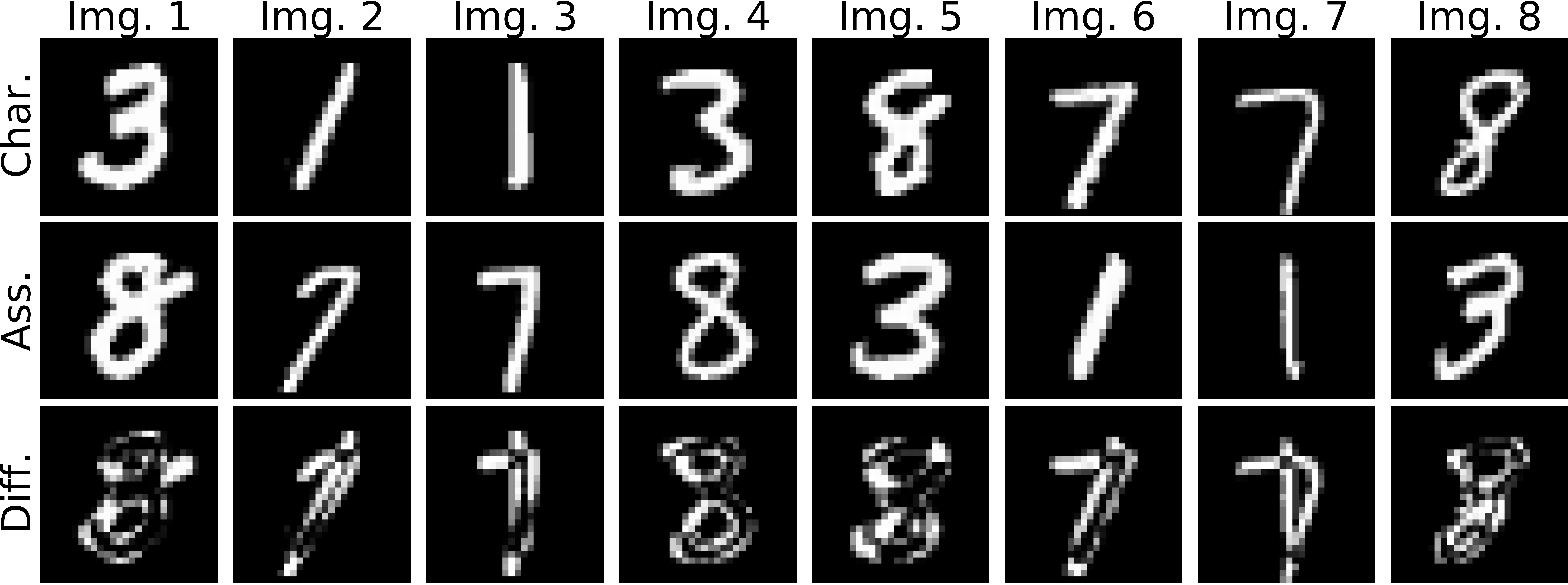}
    \subcaption{{Illustration of resulting pairs of digits: characteristic sample (Char. / top row), associated sample (Ass. / middle row) and difference (Diff. / lower row)}\label{fig:exp:mnist:detail}}
\end{minipage}
\caption{Our method applied to artificial data set created using MNIST data}
\label{fig:exp:mnist}
\end{figure}

We apply the technology ($k$-NN + $k$-means) on the electricity market  benchmark data set \cite{electricitymarketdata}, which is a  well studied drift data set \cite{DBLP:journals/corr/WebbLPG17}, and a new data set derived from MNIST \cite{mnist} by inducing drift in the occurrence of classes. To obtain an overall impression we use the dimensionality reduction technique UMAP \cite{UMAP} to project the data to the two dimensional space (Figure~\ref{fig:exp:elec:overview} and \ref{fig:exp:mnist:overview}). The color displays the identifiability. The chosen characteristic samples, as well as the associated samples are emphasized.


\paragraph*{Electricity market} The Electricity Market data set~\cite{electricitymarketdata} describes electricity pricing in South-East Australia. It records price and demand in  New South Wales and Victoria as well as the amount of power transferred between those states. All time related features have been cleaned. 
We try to explain the difference between the data set before and after the 2nd of May 1997, when a new national electricity market was introduced, which allows the trading of electricity between the states of New South Wales, Victoria, the Australian Capital Territory, and South Australia. Three new features were introduced to the data set (vicprice, vicdemand, transfer), old samples were extended by filling up with constant values. 
The data set consists of $45{,}311$ instances, with 5 features each. We randomly selected $10{,}000$ instances before and after the drift (which we consider to take place at the $17{,}423^\text{th}$ sample) to create the visualization shown in Figure~\ref{fig:exp:elec}. 

As can be seen in Figure~\ref{fig:exp:elec:detail} only the last two features (vicdemand, transfer) are relevant for drift (see Figure~\ref{fig:exp:elec:detail} Diff. -- white columns mean no drift in this feature). A further analysis showed that $\Delta \text{vicprice} \approx 0$ \cite{DBLP:journals/corr/WebbLPG17}. Furthermore it can be seen that the distribution of those attributes was extended as there exist samples after the drift comparable to those before the drift, but not the other way around (see Figure~\ref{fig:exp:elec:detail} Diff. -- only $p_\text{red} \to p_\text{blue}$ is not white).

\paragraph*{MNIST} The data set consists of sample digits 1,3,4,7,8 from the MNIST data set. The digits 1 and 3 are present before the drift, the digits 7 and 8 after the drift. 4  occurs before and after drift alike.
Each data point consists of a $28 \times 28$-pixel black-white images of numbers. We randomly selected $1{,}250$ of those images (aligned as described above)
to create the visualization shown in Figure~\ref{fig:exp:mnist}.

As can be seen in Figure~\ref{fig:exp:mnist:detail} only the digits 1,3,7,8 are considered to be relevant to the drift. The blob of data point on the left side of Figure~\ref{fig:exp:mnist:overview}, that are marked as un-identifiabe ($i = 0$), are "4"-digits, indeed. Furthermore, we observe that there is some tendency to associate "1"- and "7"-digits and "3"- and "8"-digits, as can be seen in Figure~\ref{fig:exp:mnist:overview} and \ref{fig:exp:mnist:detail}.


\section{Discussion and Further Work}
We introduced a new method to formalize an explanation of   drift observed in a distribution by means of characteristic examples, as quantified in terms of the identifiability function. We derived a new algorithm  to estimate this characteristics
and demonstrated its relation to intuitive notions of change as well as statistical notions, respectively. We demonstrated the behavior in several examples, and the  empirical results demonstrate that this proposal constitutes a promising approach as regards drift explanation in an intuitive fashion. The technology is yet restricted to discrete time points with well defined change points or points of drift. An extension to continuous drift is subject of ongoing work.

\bibliographystyle{abbrv}
\bibliography{bib}


%% file: supplement.tex
\section{Proofs}
In this section we will give complete proofs of the stated theorems. The numbering of the theorems coincide with the one given in the paper. The needed lemmas are not contained in the paper itself and follow a different numbering. 

\begin{lemma}
$i \in L^1(\P_X)$ is a well-defined measurable map and $0 \leq i(X) \leq 1$ holds $\P$-a.s..
\begin{proof}
Since $f_x(t)$ is a Radon-Nikodým density,it is a well-defined map in $L^1(\P_X)$.

Let us start by showing that $p_x$ is a probability measure, indeed. To start with notice that for $A \subset \T$ it holds
\begin{align*}
    p_x(A) &= \frac{1}{|T|} \sum_{t \in A} f_x(t) \\&= \frac{1}{|T|} \sum_{t \in A} \frac{\d p_t }{ \d \frac{1}{|\T|} \sum_{t' \in \T} p_{t'}}(x)
    \\&\overset{\textbf{!}}{=} \frac{\d \sum_{t \in A} p_t }{ \d \sum_{t \in \T} p_t},
\end{align*}
where $\textbf{!}$ holds follows by the linearity of Radon-Nikodým densities.
Furthermore for two probability measures $\mu,\nu$ we have that
\begin{align*}
    \frac{\d \mu}{\d \mu} &= 1 \\
    \frac{\d \mu}{\d \mu + \nu} & \geq 0 \\
    \Rightarrow \frac{\d \mu}{\d \mu + \nu} + \frac{\d \nu}{\d \mu + \nu} &= \frac{\d \mu + \nu}{\d \mu + \nu} = 1 \\
    \Rightarrow 0 \leq \frac{\d \mu}{\d \mu + \nu} &\leq 1
\end{align*}
where the first statement follows from the fact that, if $\mu \ll \nu$ and $\nu \ll \mu$ then $\d \mu / \d \nu = (\d \nu / \d \mu)^{-1}$ so that $\d \mu / \d \mu = (\d \mu / \d \mu)^{-1}$ and hence $\d \mu / \d \mu = 1$. 
So by writing $\sum_{t \in \T} p_t = \sum_{t \in A} p_t + \sum_{t \in A^C} p_t$ we see that $p_x$ is a probability measure on $\T$, so that we can speak of the entropy of $p_x(t)$. 

Now let us show that $i$ is measurable. Since $x \mapsto (\d p_t / \d \sum_{t' \in \T} p_{t'})(x)$ is measurable and $x \mapsto x \log x$ is measurable, as well as the sum of measurable functions is measurable it follows that $x \mapsto H(p_x)$ and hence $i$ is measurable, too.

Now, since for all probability measure $P$ on $\T$ it holds $0 \leq H(P) \leq \log |\T|$ it follows that 
\begin{align*}
     0 &= 1-\log |\T| / \log |\T| \\&\leq 1-H(p_x) / \log |\T| = i(x) \\&\leq 1-0/\log |\T| \\&= 1.
\end{align*}
\end{proof}
\end{lemma}

\begin{theorem}
It holds that $p_t$ has drift if and only if $\E[i(X)] \neq 0$.
\begin{proof}
\textbf{Suppose $p_t$ has no drift} then it holds that
\begin{align*}
f_x(t) = \frac{\d p_t}{\d \frac{1}{|\T|} \sum_{t' \in \T} p_{t'} }(x) &= \frac{\d p_t}{\d p_t \frac{1}{|\T|} \sum_{t' \in \T} 1}(x) = 1
\end{align*}
is a valid choice. Hence it follows that $p_x$ is the uniform distribution for all $x$ and hence $i(x) = 0$ since $H(\mathcal{U}(\T)) = \log |\T|$. 

\textbf{Suppose $\E[i(X)] = 0$} then $i(X) = 0$ holds $\P$-a.s. since $i \geq 0$ a.s.. Now, since for any probability measure $P$ on $\T$ it holds $H(P) = \log |\T|$ if and only if $P = \mathcal{U}(\T)$ the uniform distribution on $\T$ it follows that 
\begin{align*}
    H(p_X) &= \log |\T| & \P-a.s.
    \\\Longleftrightarrow \frac{\d p_t }{ \d\frac{1}{|\T|} \sum_{t'' \in \T} p_{t''}}(X) &= \frac{\d p_{t'} }{\d \frac{1}{|\T|} \sum_{t'' \in \T} p_{t''}}(X) & \forall t,t' \in \T \; \P-a.s.
    \\\overset{\textbf{!}^1}{\Longrightarrow} \frac{\d p_t }{ \d\frac{1}{|\T|} \sum_{t'' \in \T} p_{t''}}(X) &= \frac{\d p_{t'} }{ \d\frac{1}{|\T|} \sum_{t'' \in \T} p_{t''}}(X) & \P-a.s. \; \forall t,t' \in \T
    \\\overset{\textbf{!}^2}{\Longleftrightarrow} p_t &= p_{t'} & \forall t,t' \in \T,
\end{align*}
where $\textbf{!}^1$ follows since $\P$ is monotonous (in the second case the null sets may depend on $t,t'$) and $\textbf{!}^2$ follows from the uniqueness of Radon-Nikodým densities.%
\end{proof}
\end{theorem}

Recall the following definition:
\begin{definition}
\newcommand{\A}{\mathcal{A}}
\newcommand{\B}{\mathcal{B}}
\newcommand{\C}{\mathcal{C}}
Let $(S,\B)$ a measurable space. For a set $\C \subset \B$ we define a pseudonorm on the space of all finite measures
\begin{align*}
    \Vert P \Vert_\C = \sup_{C \in \C} P(C).
\end{align*}

Let $(\Omega,\A,\P)$ be a probability space and $X,X_1,X_2,...$ a sequence of $S$-valued, i.i.d. random variables. We say that $\C$ is a Glivenko–Cantelli class of $\P_X$ iff
\begin{align*}
    \left\Vert \frac{1}{n} \sum_{i = 1}^n \delta_{X_i} - \P_X \right\Vert_\C \to 0 \text{ a.s. }.
\end{align*}
\end{definition}

\begin{lemma}
\label{thm:av_to_mean}
\newcommand{\A}{\mathcal{A}}
\newcommand{\B}{\mathcal{B}}
\newcommand{\C}{\mathcal{C}}
Let $(\Omega,\A,\P)$ be a probability space, $(S, \B)$ a measure space and $X,X_1,X_2,...$ a sequence of $S$-valued, i.i.d. random variables. Let $f : S \to \R_{\geq 0}$ be a bounded, measurable map with $\E[f(X)] \neq 0$. Then for any set $\C \subset \B$ it holds
\begin{align*}
    \left\Vert \frac{1}{\sum_{i = 1}^n f(X_i)} \sum_{i = 1}^n \delta_{X_i} f(X_i) - \frac{1}{n \E[f(X)]} \sum_{i = 1}^n \delta_{X_i} f(X_i) \right\Vert_\C \xrightarrow{n \to \infty} 0 \text{ a.s. } ,
\end{align*}
where $\delta_x$ denotes the Dirac measure concentrated at $x$ (we use the convention $0/0 = 0$).
\begin{proof}
Denote by
\begin{align*}
    F_n &:= \frac{1}{n} \sum_{i = 1}^n f(X_i), \\
    V_n &:= \sum_{i = 1}^n \delta_{X_i} f(X_i), \\
    F &:= \E[f(X)].
\end{align*}
We hence may rewrite the statement as
\begin{align*}
    \left\Vert \frac{1}{n F_n} V_n - \frac{1}{nE}V_n \right\Vert_\C &= \frac{1}{n} \left\Vert \left(F_n^{-1} - E^{-1}\right) \cdot V_n  \right\Vert_\C \xrightarrow{n \to \infty} 0 \text{ a.s. }.
\end{align*}
Since for any $\omega \in \Omega$ we have that $F_n(\omega) > 0$ implies $F_{n+1}(\omega) > 0$ and $F_n(\omega) = 0$ implies $V_n(\omega) = 0$ we have that if there exists no $N$ such that $F_N(\omega) > 0$ we have that $V_n(\omega) \cdot (F_n^{-1}(\omega)-F) = 0$ for all $n \in \N$, on the other hand if there exists a $N$ such that $F_N(\omega) > 0$ then the sequence $F_{N+m}(\omega) > 0$ and converges to $\E[F_n] = F$ by the low of large numbers so that $F_{N+m}^{-1}(\omega)$ converges to $E^{-1}$ and hence we see that $V_n \cdot (F_n^{-1}-1) \to 0$ a.s. since $V_n(\omega) < \infty$ a.s..
\end{proof}
\end{lemma}

\begin{lemma}
\label{thm:weight_approx}
\newcommand{\A}{\mathcal{A}}
\newcommand{\B}{\mathcal{B}}
\newcommand{\C}{\mathcal{C}}
\renewcommand{\H}{\mathcal{H}}
\newcommand{\I}{\mathbb{I}}
Let $(\Omega,\A,\P)$ be a probability space, $(S, \B)$ a measure space and $X,X_1,X_2,...$ a sequence of $S$-valued, i.i.d. random variables. Let $f : S \to \R_{\geq 0}$ be a bounded, measurable map. Denote by $W(A) := \E[\I_A(X)f(X)]$ the $f$ weighted version of $\P_X$, where $\I$ denotes the indicator function. If $\C \subset \B$ is a Glivenko–Cantelli class of $\P_X$ then it holds
\begin{align*}
    \left\Vert \frac{1}{n} \sum_{i = 1}^n \delta_{X_i} f(X_i) - W \right\Vert_\C \xrightarrow{n \to \infty} 0 \text{ a.s. }.
\end{align*}
\begin{proof}
We will prove the statement using monotonous class techniques. Let $\H$ be the set of all functions $f$ such that 
\begin{align}
    \left\Vert \frac{1}{n} \sum_{i = 1}^n \delta_{X_i} f(X_i) - \E[\I_\bullet (X)f(X)] \right\Vert_\C \to 0 \text{ a.s. }.
\end{align}
Clearly $1,0 \in \H$ and by the triangle inequality it follows that if $f,g \in \H$, $\mu,\nu \in \R$ then $\mu f + \nu g \in \H$. Now, let $0 \leq f_1 \leq f_2 \leq ... \to f$ be a bounded, increasing, point wise and converging sequence with $f_m \in \H$. 
\begin{align*}
    \left\Vert \frac{1}{n} \sum_{i = 1}^n \delta_{X_i} f(X_i) - \E[\I_\bullet (X)f(X)] \right\Vert_\C 
      &\leq \quad
          \left\Vert \frac{1}{n} \sum_{i = 1}^n \delta_{X_i} f(X_i) - \frac{1}{n} \sum_{i = 1}^n \delta_{X_i} f_m(X_i) \right\Vert_\C \\&\quad +  
          \left\Vert \frac{1}{n} \sum_{i = 1}^n \delta_{X_i} f_m(X_i) - \E[\I_\bullet (X)f_m(X)] \right\Vert_\C \\&\quad + 
          \left\Vert \E[\I_\bullet (X)f_m(X)] - \E[\I_\bullet (X)f(X)] \right\Vert_\C.
\end{align*}
Since $0 \leq f$ is bounded, it is integrable for every finite measure $P$ (so in particular for $P = \P_X$ or $P = n^{-1}\sum_i \delta_{X_i}$), too. Therefore, by the dominated convergence theorem, it holds that for every $\varepsilon > 0$ we may find an $N$ such that for all $m > N$ it holds
\begin{align*}
    \sup_{A \in \C} \left| \int_A f_m \d P - \int_A f \d P \right| 
      &= \sup_{A \in \C} \left| \int_A f_m - f \d P \right| 
    \\&\leq \sup_{A \in \C} \int_A \left| f_m - f \right| \d P
    \\&\leq \int \left| f_m - f \right| \d P < \varepsilon
\end{align*}
so we see that $f \in \H$ by an $3/4 \varepsilon$-argument.
We have shown that $\H$ is an monotonous vector space, once we have shown that for any $A$ we have $\I_A \in \H$ the statement follows.

W.l.o.g. w.m.a. $\P[X \in A] > 0$. Denote by $Y_i := \I_A(X_i)$. Consider
\begin{align*}
    &\quad \P_X(A)^{-1} \left\Vert \frac{1}{n} \sum_{i = 1}^n \delta_{X_i} Y_i - \E[\I_\bullet(X) \I_A(X)] \right\Vert_\C
    \\&= \left\Vert \frac{1}{n\P_X(A)} \sum_{i = 1}^n \delta_{X_i} Y_i - \E[\I_\bullet(X) | X \in A] \right\Vert_\C
    \\&\leq \quad \frac{1}{n} \left\Vert \frac{1}{\P_X(A)} \sum_{i = 1}^n \delta_{X_i} Y_i - \frac{1}{\frac{1}{n}\sum_{i = 1}^n Y_i} \sum_{i = 1}^n \delta_{X_i} Y_i \right\Vert_\C \\&\quad+\quad \left\Vert \frac{1}{\sum_{i = 1}^n Y_i} \sum_{i = 1}^n \delta_{X_i} Y_i - \E[\I_\bullet(X) | X \in A] \right\Vert_\C.
\end{align*}
By lemma~\ref{thm:av_to_mean} we see that the first summand converges to 0 almost surly. On the other hand consider $X_i$ as an descrete stochastic process and fix its induced filtration. Define $\overline{Y}_n = \sum_{i = 1}^n Y_i$ and $\tau_n = \inf \{ i | \overline{Y}_i \geq n\}$ a sequence of stopping times, so for every fix $\omega$ we have that $X_{\tau_i}(\omega)$ is the subsequence of $X_i(\omega)$ that lies within $A$. Since $\P[\tau_i > \tau_{i-1}+n] = \P[X \in A^C]^n \to 0$ we have $\P[\tau_i = \infty] = 0$. Then $X_{\tau_i}$ is a sequence of i.i.d. random variables with distributed according to $\P_{X|X \in A}$. Since $(\sum_{i = 1}^{\tau_n} Y_i)^{-1} \sum_{i = 1}^{\tau_n} \delta_{X_i}Y_i = n^{-1} \sum_{i = 1}^n \delta_{X_{\tau_i}}$ it follows that the second summund converges to 0 almost surly. 
\end{proof}
\end{lemma}

\begin{theorem}
\newcommand{\A}{\mathcal{A}}
\newcommand{\B}{\mathcal{B}}
\newcommand{\C}{\mathcal{C}}
\renewcommand{\H}{\mathcal{H}}
\newcommand{\I}{\mathbb{I}}
\newcommand{\Xs}{\mathbf{X}}
Let $(\Omega,\A,\P)$ be a probability space, $(S, \B)$ a measure space and $X,X_1,X_2,...$ a sequence of $S$-valued, i.i.d. random variables. Let $f : S \to \R_{\geq 0}$ be a bounded, measurable map with $\E[f(X)] = 1$. Denote by $W(A) := \E[\I_A(X)f(X)]$ the $f$ weighted version of $\P_X$, where $\I$ denotes the indicator function. For every $n \in \N$ let $Y_1^{(n)},Y_2^{(n)},...$ be a sequences of independent $\{1,...,n\}$-valued random variables with $\P[Y_i^{(n)} = j|X_1,...,X_n] = f(X_j) / \sum_{k = 1}^n f(X_k)$ (or $1/n$ iff all $f(X_i) = 0$) for all $i = 1,2,...$ and $j = 1,...,n$. If $\C \subset \B$ is a Glivenko–Cantelli class of $\P_X$ then it holds
\begin{align*}
    W_{n,m} := \frac{1}{m} \sum_{i = 1}^m \delta_{X_{Y_i^{(n)}}} &\xrightarrow{m,n \to \infty} W 
\end{align*}
in $\Vert \cdot \Vert_\C$ almost surely, where we can take the limits in arbitrary order. 
\begin{proof}
Denote by $W_n := \E[W_{n,m}|X_1,...,X_n]$ the theoretical measure for a fixed set of observations and by $\tilde{W}_n = n^{-1}\sum_{i = 1}^n \delta_{X_i}f(X_i)$. It holds
\begin{align*}
    \Vert W_{n,m} - W \Vert_\C &\leq \Vert W_{n,m} - W_n \Vert_\C + \Vert W_n - \tilde{W}_n \Vert_\C + \Vert \tilde{W}_n - W \Vert_\C.
\end{align*}
Notice that only the first summand depend on $m$ and $n$. However, since we approximate an distribution on $\{1,...,n\} \subset \R$ we see that by Kolmogorov's theorem $\Vert W_{n,m} - W_n \Vert_\C$ is uniformly bounded by $m$.

By lemma~\ref{thm:weight_approx} we see that $\Vert \tilde{W}_n - W \Vert_\C \to 0$ a.s. and hence it remains to show that $\Vert W_n - \tilde{W}_n \Vert_\C \to 0$ a.s.: Denote by $\Xs = X_1,...,X_n$, $A_i = \{Y_1^{(n)} = i\}$. We have that
\begin{align*}
    W_n(A) 
    &= \E\left[\frac{1}{m}\sum_{j = 1}^m \I_A(X_{Y_i^{(n)}}) | \Xs \right]
  = \E\left[\I_A(X_{Y_1^{(n)}}) | \Xs \right]
  \\&= \E\left[\sum_{i = 1}^n \I_A(X_i) \I_{A_i} | \Xs \right]
  = \sum_{i = 1}^n \I_A(X_i) \E\left[ \I_{A_i} | \Xs \right]
  \\&= \sum_{i = 1}^n \I_A(X_i) \P\left( A_i | \Xs \right)
  = \sum_{i = 1}^n \I_A(X_i) \frac{f(X_i)}{\sum_{j = 1}^n f(X_j)} & \text{Def. $A_i$ and $Y_i^{(n)}$}
  \\&= \frac{1}{n} n \sum_{i = 1}^n \I_A(X_i) f(X_i) \left(\sum_{j = 1}^n f(X_j)\right)^{-1}
  \\&= \left( \frac{1}{n} \sum_{i = 1}^n \I_A(X_i) f(X_i) \right) \cdot \left(\frac{1}{n}\sum_{i = 1}^n f(X_i)\right)^{-1},
\end{align*}
so the statement follows by lemma~\ref{thm:av_to_mean}.
\end{proof}
\end{theorem}

\section{Experiments}
In this section we will give additional details on the evaluations and experiments. This includes a precise setup of how the used data was generated and how the experiments were run as well as the obtained results/measurements and our interpretations. 

\subsection{Experimental setup}
In this subsection we will discuss how we generated our data and how we evaluated the results. To simplify it we use different paragraphs for theoretical and benchmark data.
\paragraph*{Theoretical data}
As discussed in the paper we were interested in understanding which of the following parameters is relevant for the quality of our prediction:
\begin{enumerate}
    \item Dimension (Dimensionality of data)
    \item Complexity of distribution (How complex/fractal/fine grained is $p_t$)
    \item Complexity of overlap (How complex/fractal/fine grained are the regions where $p_t$ and $p_{t'}$ have weight)
\end{enumerate}
We used a mixture of Gaussians with uniformly distributed means and constant variance. We controlled the dimensionality in the obvious way ($d$). We controlled the complexity of the distributions by the number of used Gaussians with equal degree of overlap (${n_\text{Gauss per Cluss}}$). We controlled the complexity of overlap by controlling the number of degrees of overlap ($n_\text{Class}$). 

We therefore obtain $\mu_{i,j} \sim \mathcal{U}([-a,a]^d)$ and
\begin{align*}
    p_{d,{n_\text{Cluss}},{n_\text{Gauss per Class}}}(x,t) &= \sum_{i = 1}^{n_\text{Gauss per Cluss}} \sum_{j = 1}^{n_\text{Class}} \mathcal{N}^d(\mu_{ij},\sigma) \times \left(\frac{j}{{n_\text{Class}}}\delta_1+\left(1-\frac{j}{{n_\text{Class}}}\right)\delta_2\right).
\end{align*}
Notice that $p_{d,{n_\text{Cluss}},{n_\text{Gauss per Class}}}$ is a distribution on $\R^d \times \T$ with $\T = \{1,2\}$.
In this case $i$ and $C$ can be computed analytically given $p_{d,{n_\text{Cluss}},{n_\text{Gauss per Class}}}$.

We generated 500 samples according to $p_{d,{n_\text{Cluss}},{n_\text{Gauss per Class}}}$. 

For the evaluation of the estimation of $i$ we trained our models on the data to solve the probabilistic classification task $h : \R^d \to \text{Prob}(\T), x \mapsto (p_1,p_2)$, i.e. the classification rule for a sample $x$ is given by $\argmax_{t \in \T} h(x)_t$. We evaluated the resulting models by estimating the MSE between the estimation $i_h$ (based on $h$) and the real $i$ using 1.500 samples distributed according to $p_{d,{n_\text{Cluss}},{n_\text{Gauss per Class}}}$, 1.500 samples distributed according to a Gaussian mixture equivalent to $p_{d,{n_\text{Cluss}},m}$ except that we used $3\sigma$ and 1.500 samples distributed according to $\mathcal{U}([-a,a]^d)$. We repeated the process for every considered combination of $d,{n_\text{Cluss}},{n_\text{Gauss per Class}}$ 30 times and document mean value and standard deviation.

Notice that the classifier is not trained on data which contains $i(x)$! Instead it is trained to predict the time point $t$ given $x$. Since we consider probabilistic models this allows us to use them to estimate $i_h$, but the actual value of $i(x)$ is never presented to the model. 

For the evaluation of the estimation of $C$ we applied the (modified) clustering methods to the generated samples. The ground truth value of $i$ was used by the methods. We evaluated the resulting models using the ground truth value of $C$ and $i$. If a clustering produced more then one prototype the mean value over all prototypes was considered as the accomplish value for the maximization of $C$ and $i$. We repeated the process for every considered combination of $d,{n_\text{Cluss}},{n_\text{Gauss per Class}}$ 30 times and document mean value and standard deviation.

\paragraph*{Benchmark data}
\newcommand{\C}{\mathcal{C}}
We considered both regression data ($\D \subset \R^d \times \R$) and classification data ($\D \subset \R^d \times \C$). 
We processed the regression data as follows: We normalized the data sets output, i.e. we have $\D \subset \R^d \times [0,1]$. For every sample $(x,y) \in \D$ we randomly generated a occurrence time $t \in \{1,2\}$ with $t \sim \text{Ber}(y)$, i.e. the chance that $t = 1$ is higher if the original prediction value $y$ is close to 0. Accordingly we computed the identifiability as $i = 1-H(\text{Ber}(y))/\log 2$. The new samples are given by $(x,t,i)$.

We processed the classification data as follows: For every class $y \in \C$ we randomly choose a fix occurrence probability $p_y \in [0,1]$. For every sample $(x,y) \in \D$ we randomly generated a occurrence time $t \in \{1,2\}$ with $t \sim \text{Ber}(p_y)$, i.e. some classes are more likely to occur before resp. after the drift. Accordingly we computed the identifiability as $i = 1-H(\text{Ber}(y_c))/\log 2$. The new samples are given by $(x,t,i)$.

We split our data sets randomly into test and training set (50\%/50\%). 

For the evaluation of the estimation of $i$, we trained our models on the training set to solve the probabilistic classification task $h : \R^d \to \text{Prob}(\T), x \mapsto (p_1,p_2)$, i.e. the classification rule for a sample $x$ is given by $\argmax_{t \in \T} h(x)_t$. We evaluated the resulting models by estimating the MSE between the estimation $i_h$ (based on $h$) and the "real" $i$ defined when preprocessing the data set. We repeated the process for every data set 30 times and document mean value and standard deviation.

Notice that the classifier is not trained on data which contains $i$! Instead it is trained to predict the time point $t$ given $x$. Since we consider probabilistic models this allows us to use them to estimate $i_h$, but the actual value of $i$ is never presented to the classifier. 

For the evaluation of the estimation of $C$ we applied the (modified) clustering methods to the generated samples. The defined value of $i$ was used by the methods. We evaluated the resulting models using the defined value of $i$. If a clustering produced more then one prototype the mean value over all prototypes was considered as the accomplish value for the maximization of $i$. Since the estimation of $C$ was too unstable and it seemed to us, when considering the theoretical results, that no further befit would result from it, we omitted it. We repeated the process for every data set 30 times and document mean value and standard deviation.

\paragraph*{Hyperparameters} In any case we used standard parameters. 

\paragraph*{Data sets} We used the following data sets:
\begin{itemize}
    \item electricity: \url{http://moa.cms.waikato.ac.nz/datasets/} (Normalized Dataset) 
    \item MNIST: \url{https://www.openml.org/d/554}
    \item digits: \url{https://archive.ics.uci.edu/ml/datasets/Optical+Recognition+of+Handwritten+Digits} (test set only)
    \item (breast) cancer: \url{https://goo.gl/U2Uwz2}
    \item wine: \url{https://archive.ics.uci.edu/ml/machine-learning-databases/wine/wine.data}
    \item iris: \url{https://www.openml.org/d/61}
    \item diabetes: \url{https://www.openml.org/d/41514}
    \item boston: \url{https://www.openml.org/d/531}
    \item (olivetti) faces: \url{https://www.openml.org/d/41083}
\end{itemize}
Further characteristics are presented in table~\ref{tab:datasets}.
\begin{table}[]
    \centering
    \caption{Data set characteristics.}
    \begin{tabular}{ccc}
        \hline
        data set & samples & features \\
        \hline
        electricity & 45312 & 7 \\
        MNIST & 70000 & 784 \\
        digits & 1797 & 64\\
        cancer & 569 & 30\\
        wine & 178 & 13\\
        iris & 150 & 4\\
        diabetes & 442 & 10\\
        boston & 506 & 13\\
        faces & 400 & 4096\\
        \hline
    \end{tabular}
    \label{tab:datasets}
\end{table}

\subsection{Results}
We will now present and discuss our results.

\begin{figure}[!tb]
    \centering
    \includegraphics[width=\textwidth]{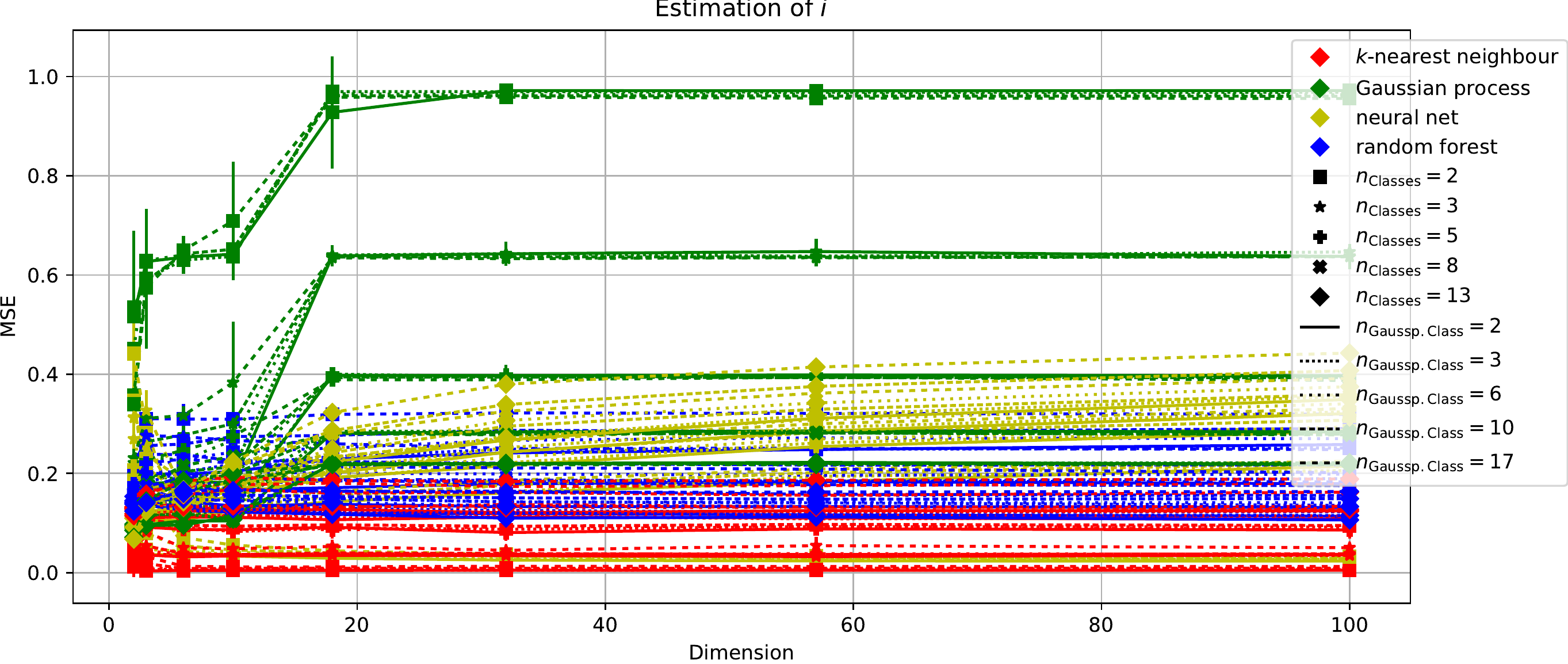}
    \caption{Evaluation of different models on theoretical data. Graphic shows MSE with respect to ground truth. Color represents used model ($k$-nearest neighbour classifier ($k$-NN, $k = 5$), Gaussian process classifier (GP, Matern-kernel), artificial neural network (ANN, 1-hidden layer with 100 neurons) and random forest (RF, 10 random trees)), marker degrees of overlap, line style complexity of distribution. Estimation over 30 runs.}
    \label{fig:all_ident}
\end{figure}

\paragraph*{Evaluation of estimation of $i$ on theoretical data with known ground truth} We performed the evaluation described above, the results are presented in Figure~\ref{fig:all_ident}. As can be seen $k$-NN performs best over all configurations, follows by random forest.  Gaussian process fails in particular when there is no overlap, this is also true for all methods except $k$-NN, where we observe the opposite. GP and neuronal network suffers from problems when facing an increasing dimension. However, other then ANN GP seems to stabilize for large dimension. 

\begin{figure}[!tb]
    \centering
    \includegraphics[width=\textwidth]{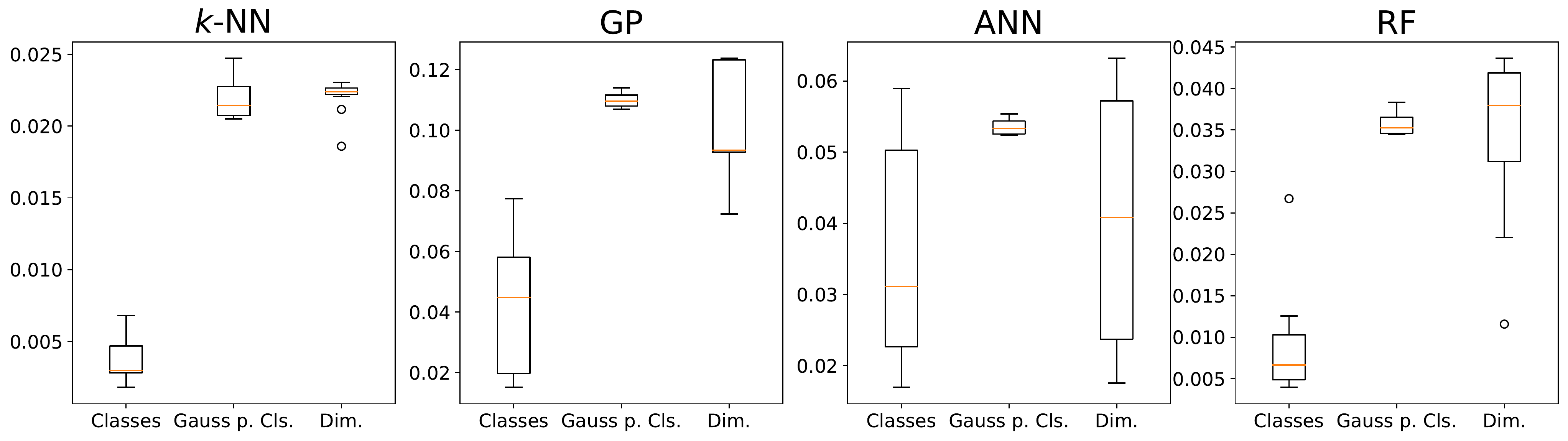}
    \caption{Effect strength on MSE of estimation of $i$ with respect to ground truth via conditional variance of parameters for different models. $k$-nearest neighbour ($k$-NN), Gaussian process (GP), artificial neural network (ANN) and random forest (RF). Estimation over 30 runs. Based on data presented on Figure~\ref{fig:all_ident}.}
    \label{fig:effect_strength_ident}
\end{figure}

We also evaluated the impact of our parameters $d,{n_\text{Gauss per Cluss}}$ and ${n_\text{Cluss}}$. We present an estimation of important quantities of the distribution of the conditional variation of our observation given the respective parameters in Figure~\ref{fig:effect_strength_ident}. As we use conditional variation a smaller value implies a larger impact, since it allows us to predict the resulting value with small error. As can be seen the complexity of overlap is by far the most important parameter, followed by dimensionality. The complexity of the distributions seems not all to relevant. However, this may be caused by the used distribution.

\begin{table}[!tb]
    \centering
    \caption{MSE for estimation of the identifiability function using different models ($k$-nearest neighbour classifier ($k$-NN, $k = 5$), Gaussian process classifier (GP, Matern-kernel), artificial neural network (ANN, 1-hidden layer with 100 neurons) and random forest (RF, 10 random trees)). Estimation over 30 runs. Standard deviation is only shown if $\geq 0.01$.}
    \begin{tabular}{ccccc}
        data set &kNN&GP&ANN&RF\\ 
        \hline
digits&$0.12(\pm 0.01)$&$0.23(\pm 0.07)$&$0.19(\pm 0.02)$&$0.12(\pm 0.02)$\\
cancer&$0.13(\pm 0.02)$&$0.25(\pm 0.06)$&$0.13(\pm 0.03)$&$0.16(\pm 0.02)$\\
iris&$0.11(\pm 0.05)$&$0.16(\pm 0.05)$&$0.35(\pm 0.19)$&$0.12(\pm 0.04)$\\
wine&$0.32(\pm 0.10)$&$0.43(\pm 0.17)$&$0.28(\pm 0.07)$&$0.20(\pm 0.04)$\\
boston&$0.14(\pm 0.02)$&$0.06(\pm 0.01)$&$0.09(\pm 0.03)$&$0.11(\pm 0.02)$\\
diabetes&$0.13(\pm 0.02)$&$0.06(\pm 0.01)$&$0.06(\pm 0.01)$&$0.10(\pm 0.01)$\\
faces&$0.16(\pm 0.02)$&$0.16(\pm 0.02)$&$0.16(\pm 0.02)$&$0.14(\pm 0.02)$\\
    \hline
    \end{tabular}
    \label{tab:est_i}
\end{table}

\paragraph*{Evaluation of estimation of $i$ on benchmark data with unknown ground truth} We performed the evaluation described above, the results are presented in Table~\ref{tab:est_i}. Except for the wine data set $k$-NN and RF perform very comparable and better then GP on most of the data sets. The only data sets where GP performs better are regression data sets and the faces data set which has about 40 classes, i.e. those are data set with very complicated overlap. This matches our findings from the theoretical data. 

\begin{figure}[!tb]
    \centering
    \includegraphics[width=\textwidth]{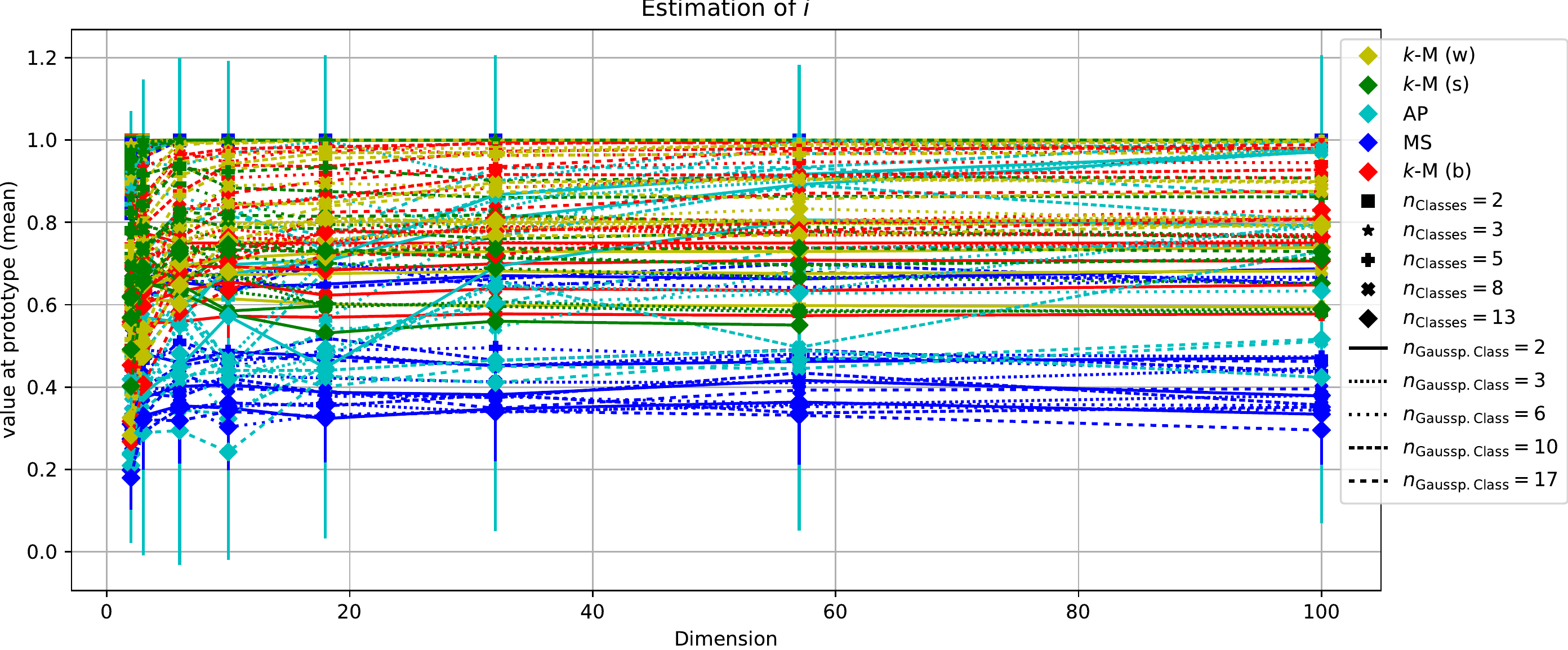}
    \caption{Evaluation of different models on theoretical data: Finding identifiable points in data, (mean) value of $i$ at found prototype(s). Using $k$-Means with weighting and resampling (sampled) as well as without any further consideration as baseline, Affinity Propagation and Mean Shift (both where applied with resampling). Evaluation over 30 runs. Color represents used model, marker degrees of overlap, line style complexity of distribution.}
    \label{fig:proto_ident}
\end{figure}

\begin{figure}[!tb]
    \centering
    \includegraphics[width=\textwidth]{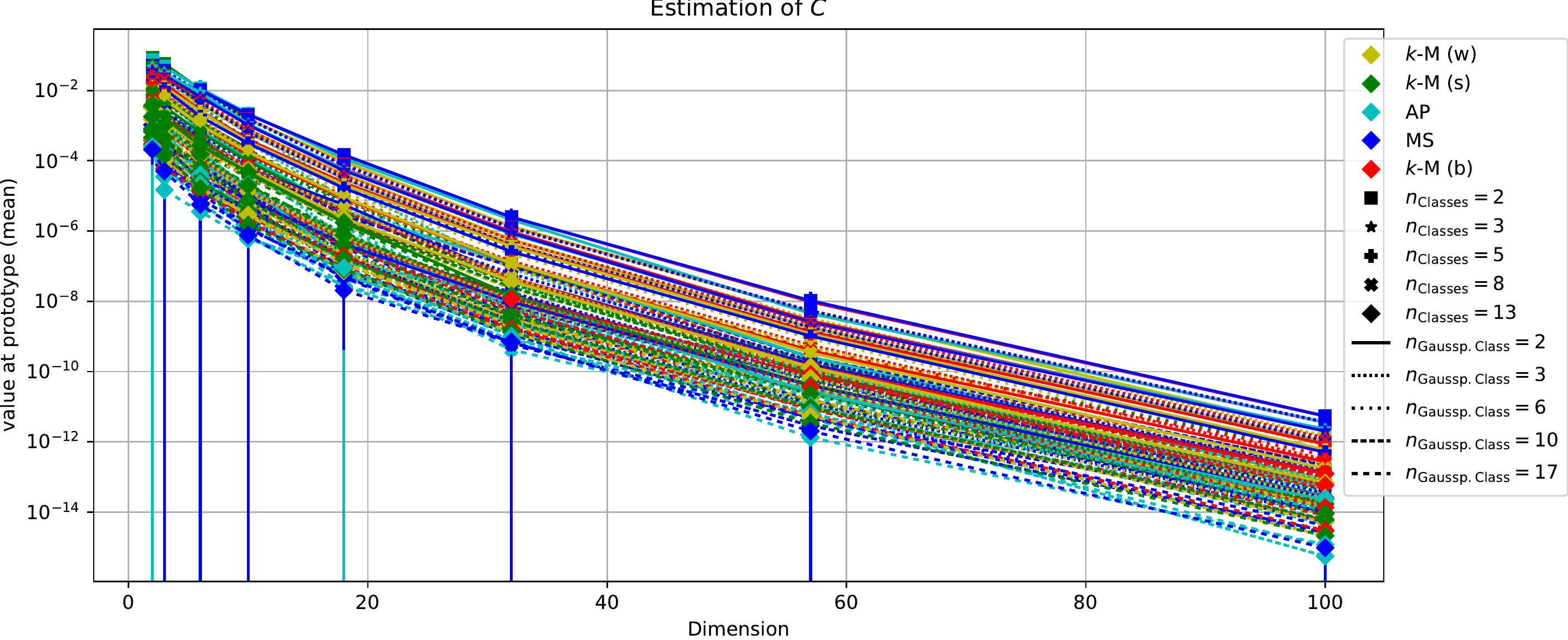}
    \caption{Evaluation of different models on theoretical data: Finding characteristic points in data, (mean) value of $C$ at found prototype(s). Using $k$-Means with weighting and resampling (sampled) as well as without any further consideration as baseline, Affinity Propagation and Mean Shift (both where applied with resampling). Evaluation over 30 runs. Color represents used model, marker degrees of overlap, line style complexity of distribution.}
    \label{fig:proto_char}
\end{figure}

\begin{figure}[!tb]
    \centering
    \includegraphics[width=\textwidth]{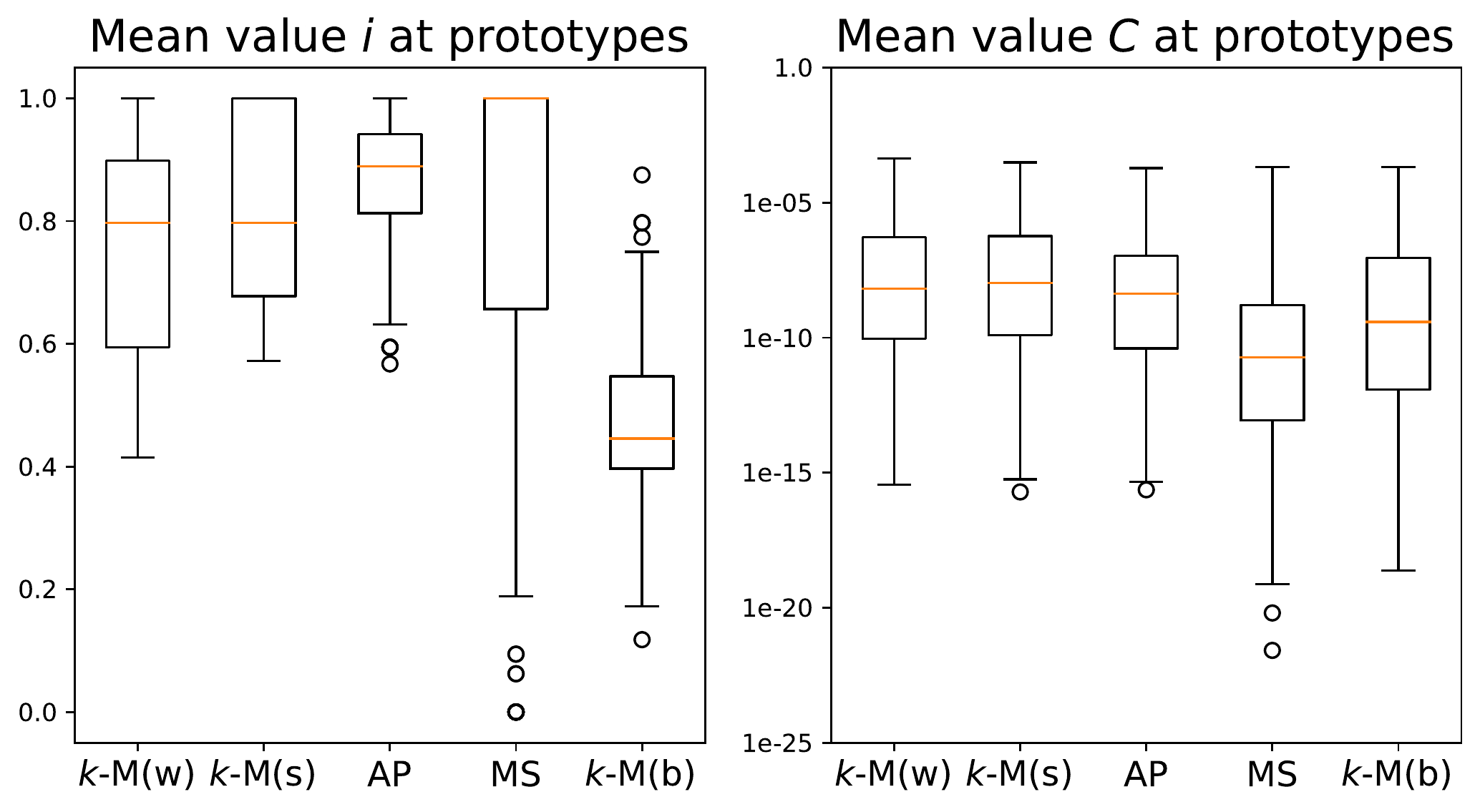}
    \caption{Condensed comparison of different models on theoretical data: Finding identifiable resp. characteristic points in data, (mean) value of $i$ resp. $C$ at found prototype(s). Using $k$-Means with weighting and resampling as well as without any further consideration as baseline ($k$-M (w) / (s) / (b)), Affinity Propagation (AP) and Mean Shift (MS). AP and MS where applied with resampling. Estimation over 30 runs. Based on data presented on Figures~\ref{fig:proto_ident} and \ref{fig:proto_char}.}
    \label{fig:proto_box}
\end{figure}

\paragraph*{Evaluation of maximization of $i$ and $C$ on theoretical data with known ground truth} We performed the evaluation described above, the results are presented in the following figures: Figure~\ref{fig:proto_ident} maximization of $i$ using different clustering algorithms and parameters of distribution; Figure~\ref{fig:proto_char} maximization of $C$ using different clustering algorithms and parameters of distribution; Figure~\ref{fig:proto_box} summery of the maximization of $i$ and $C$ using different clustering algorithms.

As can be seen in Figure~\ref{fig:proto_ident} there seems to be no real impact of the distribution parameters when it comes to optimizing $i$.

As can be seen in Figure~\ref{fig:proto_char} there seems to be no real impact of the distribution parameters except dimension $d$ when it comes to optimizing $C$. This effect may be due to the chosen distribution since normal distributions are known to suffer from the curse of dimensionality.

As can be seen in Figure~\ref{fig:proto_box} resampling seems to be a reasonable approach when it comes to optimizing $i$, this becomes very obvious when comparing $k$-means with resampling and weighting. It is also worth noting that all methods perform better then the baseline method. Mean shift produces some variance and outlier, though. 

When it comes to optimizing $C$, $k$-means with resampling and weighting perform similar and better then all others, followed by affinity propagation. It is worth noting that the baseline method, which does not take $i$ into account, performs comparably good, and even better then mean shift. This may imply that the error term due to the density of $X$ may outrank the error term due to $i$. 

All in all it seems not profitable to consider $C$ directly for the evaluation of methods. 

\begin{table}[!tb]
\begin{small}
    \centering
    \caption{Evaluation of method on benchmark data. Estimation of identifiability at found prototypes, value of $i$ at found prototype(s). If more then one prototype was found then mean value was used (in each run). Estimation over 30 runs. }
    \begin{tabular}{cccccccc}
method & digits & wine & boston & diabetes & faces \\
\hline
k-Means (weighted) & $0.68(\pm 0.10)$ & $0.68(\pm 0.07)$ & $0.61(\pm 0.06)$ & $0.67(\pm 0.11)$ & $0.76(\pm 0.07)$\\
k-Means (sampled) & $0.84(\pm 0.07)$ & $0.78(\pm 0.08)$ & $0.66(\pm 0.08)$ & $0.74(\pm 0.09)$ & $0.80(\pm 0.10)$\\
Affinity Propagation & $0.87(\pm 0.05)$ & $0.78(\pm 0.12)$ & $0.64(\pm 0.16)$ & $0.77(\pm 0.14)$ & $0.82(\pm 0.10)$\\
Mean Shift & $0.66(\pm 0.36)$ & $0.72(\pm 0.12)$ & $0.49(\pm 0.16)$ & $0.40(\pm 0.27)$ & $0.34(\pm 0.18)$\\
k-Means (simple) & $0.49(\pm 0.07)$ & $0.55(\pm 0.06)$ & $0.32(\pm 0.07)$ & $0.27(\pm 0.09)$ & $0.29(\pm 0.11)$\\
    \hline
    \end{tabular}
    \label{tab:eval_C}
    \end{small}
\end{table}

\paragraph*{Evaluation of maximization of $i$ and $C$ on benchmark data with unknown ground truth} We performed the evaluation described above, the results are presented in Table~\ref{tab:eval_C}. As can be seen Affinity propagation performs best over all data sets followed by $k$-means with resampling. Mean shift teds to have large variation. All methods outperform the base line. This matches our findings from the theoretical data. 


%% file: main.bbl
\begin{thebibliography}{10}

\bibitem{Aminikhanghahi:2017:SMT:3086013.3086037}
S.~Aminikhanghahi and D.~J. Cook.
\newblock A survey of methods for time series change point detection.
\newblock {\em Knowl. Inf. Syst.}, 51(2):339--367, May 2017.

\bibitem{artelt2020convex}
A.~Artelt and B.~Hammer.
\newblock Convex density constraints for computing plausible counterfactual
  explanations, 2020.

\bibitem{eddm}
M.~Baena-García, J.~Campo-Ávila, R.~Fidalgo-Merino, A.~Bifet, R.~Gavald, and
  R.~Morales-Bueno.
\newblock Early drift detection method.
\newblock 01 2006.

\bibitem{adwin}
A.~Bifet and R.~Gavald{\`{a}}.
\newblock Learning from time-changing data with adaptive windowing.
\newblock In {\em Proceedings of the Seventh {SIAM} International Conference on
  Data Mining, April 26-28, 2007, Minneapolis, Minnesota, {USA}}, pages
  443--448, 2007.

\bibitem{Bish2}
C.~M. Bishop.
\newblock {\em Pattern Recognition and Machine Learning (Information Science
  and Statistics)}.
\newblock Springer-Verlag, Berlin, Heidelberg, 2006.

\bibitem{LSDD}
L.~{Bu}, C.~{Alippi}, and D.~{Zhao}.
\newblock A pdf-free change detection test based on density difference
  estimation.
\newblock {\em IEEE Transactions on Neural Networks and Learning Systems},
  29(2):324--334, Feb 2018.

\bibitem{ijcai2019-876}
R.~M.~J. Byrne.
\newblock Counterfactuals in explainable artificial intelligence (xai):
  Evidence from human reasoning.
\newblock In {\em Proceedings of the Twenty-Eighth International Joint
  Conference on Artificial Intelligence, {IJCAI-19}}, pages 6276--6282.
  International Joint Conferences on Artificial Intelligence Organization, 7
  2019.

\bibitem{DBLP:journals/eswa/MelloVFB19}
R.~F. de~Mello, Y.~Vaz, C.~H.~G. Ferreira, and A.~Bifet.
\newblock On learning guarantees to unsupervised concept drift detection on
  data streams.
\newblock {\em Expert Syst. Appl.}, 117:90--102, 2019.

\bibitem{hdddm}
G.~Ditzler and R.~Polikar.
\newblock Hellinger distance based drift detection for nonstationary
  environments.
\newblock In {\em 2011 {IEEE} Symposium on Computational Intelligence in
  Dynamic and Uncertain Environments, {CIDUE} 2011, Paris, France, April 13,
  2011}, pages 41--48, 2011.

\bibitem{DBLP:journals/cim/DitzlerRAP15}
G.~Ditzler, M.~Roveri, C.~Alippi, and R.~Polikar.
\newblock Learning in nonstationary environments: {A} survey.
\newblock {\em {IEEE} Comp. Int. Mag.}, 10(4):12--25, 2015.

\bibitem{NIPS2019_9511}
A.-K. Dombrowski, M.~Alber, C.~Anders, M.~Ackermann, K.-R. M\"{u}ller, and
  P.~Kessel.
\newblock Explanations can be manipulated and geometry is to blame.
\newblock In H.~Wallach, H.~Larochelle, A.~Beygelzimer, F.~d\textquotesingle
  Alch\'{e}-Buc, E.~Fox, and R.~Garnett, editors, {\em Advances in Neural
  Information Processing Systems 32}, pages 13589--13600. Curran Associates,
  Inc., 2019.

\bibitem{MeanShift}
K.~Fukunaga and L.~D. Hostetler.
\newblock The estimation of the gradient of a density function, with
  applications in pattern recognition.
\newblock {\em IEEE Trans. Inf. Theory}, 21:32--40, 1975.

\bibitem{ddm}
J.~Gama, P.~Medas, G.~Castillo, and P.~P. Rodrigues.
\newblock Learning with drift detection.
\newblock In {\em Advances in Artificial Intelligence - {SBIA} 2004, 17th
  Brazilian Symposium on Artificial Intelligence, S{\~{a}}o Luis,
  Maranh{\~{a}}o, Brazil, September 29 - October 1, 2004, Proceedings}, pages
  286--295, 2004.

\bibitem{asurveyonconceptdriftadaption}
J.~a. Gama, I.~\v{Z}liobait\.{e}, A.~Bifet, M.~Pechenizkiy, and A.~Bouchachia.
\newblock A survey on concept drift adaptation.
\newblock {\em ACM Comput. Surv.}, 46(4):44:1--44:37, Mar. 2014.

\bibitem{DBLP:conf/dsaa/GilpinBYBSK18}
L.~H. Gilpin, D.~Bau, B.~Z. Yuan, A.~Bajwa, M.~Specter, and L.~Kagal.
\newblock Explaining explanations: An overview of interpretability of machine
  learning.
\newblock In F.~Bonchi, F.~J. Provost, T.~Eliassi{-}Rad, W.~Wang, C.~Cattuto,
  and R.~Ghani, editors, {\em 5th {IEEE} International Conference on Data
  Science and Advanced Analytics, {DSAA} 2018, Turin, Italy, October 1-3,
  2018}, pages 80--89. {IEEE}, 2018.

\bibitem{DBLP:journals/kais/GoldenbergW19}
I.~Goldenberg and G.~I. Webb.
\newblock Survey of distance measures for quantifying concept drift and shift
  in numeric data.
\newblock {\em Knowl. Inf. Syst.}, 60(2):591--615, 2019.

\bibitem{5376}
A.~Gretton, A.~Smola, J.~Huang, M.~Schmittfull, K.~Borgwardt, and
  B.~Sch{\"o}lkopf.
\newblock {\em Covariate shift and local learning by distribution matching},
  pages 131--160.
\newblock MIT Press, Cambridge, MA, USA, 2009.

\bibitem{Gunningeaay7120}
D.~Gunning, M.~Stefik, J.~Choi, T.~Miller, S.~Stumpf, and G.-Z. Yang.
\newblock Xai{\textemdash}explainable artificial intelligence.
\newblock {\em Science Robotics}, 4(37), 2019.

\bibitem{electricitymarketdata}
M.~Harries, U.~N. cse tr, and N.~S. Wales.
\newblock Splice-2 comparative evaluation: Electricity pricing.
\newblock Technical report, 1999.

\bibitem{mnist}
Y.~LeCun and C.~Cortes.
\newblock {MNIST} handwritten digit database.
\newblock 2010.

\bibitem{looveren2019interpretable}
A.~V. Looveren and J.~Klaise.
\newblock Interpretable counterfactual explanations guided by prototypes, 2019.

\bibitem{DBLP:journals/kais/LosingHW18}
V.~Losing, B.~Hammer, and H.~Wersing.
\newblock Tackling heterogeneous concept drift with the self-adjusting memory
  {(SAM)}.
\newblock {\em Knowl. Inf. Syst.}, 54(1):171--201, 2018.

\bibitem{UMAP}
L.~McInnes, J.~Healy, and J.~Melville.
\newblock Umap: Uniform manifold approximation and projection for dimension
  reduction, 2018.

\bibitem{molnar2019interpretable}
C.~Molnar.
\newblock Interpretable machine learning. 2019.
\newblock {\em URL [https://christophm. github. io/interpretable-ml-book/].
  accessed}, pages 05--04, 2019.

\bibitem{JMLR:v19:18-251}
J.~Montiel, J.~Read, A.~Bifet, and T.~Abdessalem.
\newblock Scikit-multiflow: A multi-output streaming framework.
\newblock {\em Journal of Machine Learning Research}, 19(72):1--5, 2018.

\bibitem{pagehinkley}
E.~S. PAGE.
\newblock {Continuous inspection schemes}.
\newblock {\em Biometrika}, 41(1-2):100--115, 06 1954.

\bibitem{Wachter2017CounterfactualEW}
S.~Wachter, B.~D. Mittelstadt, and C.~Russell.
\newblock Counterfactual explanations without opening the black box: Automated
  decisions and the gdpr.
\newblock {\em ArXiv}, abs/1711.00399, 2017.

\bibitem{Wald}
A.~Wald.
\newblock Sequential tests of statistical hypotheses.
\newblock {\em The Annals of Mathematical Statistics}, 16(2):117--186, 1945.

\bibitem{DBLP:journals/connection/WangMCY19}
S.~Wang, L.~L. Minku, N.~V. Chawla, and X.~Yao.
\newblock Learning from data streams and class imbalance.
\newblock {\em Connect. Sci.}, 31(2):103--104, 2019.

\bibitem{DBLP:journals/corr/WebbLPG17}
G.~I. Webb, L.~K. Lee, F.~Petitjean, and B.~Goethals.
\newblock Understanding concept drift.
\newblock {\em CoRR}, abs/1704.00362, 2017.

\bibitem{DBLP:journals/tnn/ZambonAL18}
D.~Zambon, C.~Alippi, and L.~Livi.
\newblock Concept drift and anomaly detection in graph streams.
\newblock {\em {IEEE} Trans. Neural Networks Learn. Syst.}, 29(11):5592--5605,
  2018.

\end{thebibliography}
